\documentclass[runningheads, envcountsame, a4paper]{llncs}

\usepackage{microtype}
\usepackage[pdftex]{graphicx}
\usepackage{epstopdf}
\usepackage{subfig}
\usepackage[hyphens]{url}
\usepackage{microtype}

\usepackage{booktabs} 

\usepackage{amsmath}
\usepackage{amssymb}
\usepackage{adjustbox}

\usepackage{multirow}
\usepackage[inline]{enumitem}
\usepackage{array}

\usepackage{xcolor}

\usepackage{hyperref}

\newcommand{\norm}[1]{\left\lVert#1\right\rVert}

\begin{document}

\title{Ensembling Shift Detectors: \\an Extensive Empirical Evaluation}

\titlerunning{Ensembling Shift Detectors}


\author{Simona Maggio*\inst{1}\orcidID{0000-0001-5653-937X} \\
Léo Dreyfus-Schmidt*\inst{2}\orcidID{0000-0001-8271-1217}}

\authorrunning{S. Maggio and L. Dreyfus-Schmidt}

\institute{Dataiku Lab, Paris, France \email{simona.maggio@dataiku.com} \and
Dataiku Lab, Paris, France \email{leo.dreyfus-schmidt@dataiku.com}\\
* Equal contribution}

\tocauthor{Simona~Maggio and Léo~Dreyfus-Schmidt}
\toctitle{Ensembling Shift Detectors: an Extensive Empirical Evaluation}

\maketitle
\setcounter{footnote}{0}

\begin{abstract}
The term \textit{dataset shift} refers to the situation where the data used to train a machine learning model is different from where the model operates.
While several types of shifts naturally occur, existing shift detectors are usually designed to address only a specific type of shift. 
We propose a simple yet powerful technique to ensemble complementary shift detectors, while tuning the significance level of each detector's statistical test to the dataset. This enables a more robust shift detection, capable of addressing all different types of shift, which is essential in real-life settings where the precise shift type is often unknown. 
This approach is validated by a large-scale statistically sound benchmark study over various synthetic shifts applied to real-world structured datasets. 


\end{abstract}

\section{Introduction}
\label{intro}

It is crucial for ML practitioners to detect possibly harmful changes of the data consumed by ML models, thus identifying situations of \textit{dataset shift}, where the joint distribution $\mathbb{P}(X, Y)$ of the input features $X$ and output $Y$ in the source domain $S$, used to train the model, is different from the distribution in the target domain $T$, where the model operates. 

There are many reasons why target domain data is different in practice from carefully collected source data. For instance, by sample selection bias, when some samples are less likely to be included in the source datasets because of a specific selection process implicitly dependent on the output variable. This is often the case for data collected through web surveys, where self-selection and under-representation of people with limited internet access prevent reliable inference \cite{Bethlehem2010}. Another example of dataset shift affecting the model validity is when the target variable distribution is not stationary. This is the case for pneumonia diagnosis models, which are sensitive to seasonal outbreaks \cite{lipton18}. 

Dataset shifts are usually categorised as follows: 
\begin{enumerate*}[label=\itshape(\roman*), itemjoin={{, }}, itemjoin*={{, while }}]
  \item prior shift indicates that the only changing factor is the prior distribution $\mathbb{P}_S(Y)\neq \mathbb{P}_T(Y)$
  \item covariate shift indicates that the only changing factor is the covariates distribution  $\mathbb{P}_S(X) \neq \mathbb{P}_T(X)$
  \item sample selection bias impacts both prior and covariate distributions.
\end{enumerate*} 
Another cause for dataset change is concept shift, where the dependence of the target variable on the features changes from source to target domain. In this work we focus on unsupervised approaches, while concept shift detection requires target labels and so is not covered in this study. 

To protect deployed models from dataset shift, we need to compare source and target data. The most challenging aspect of shift detection stems from the absence of ground truth for the target variable, preventing the use of simple model metric monitoring techniques. Thus, in order to detect changes in the output distribution, it is essential to find robust proxies. 

There are two dominant approaches for generic shift detection, one comparing feature distributions, either directly or through a domain-discriminating classifier trained on the features from the two domains \cite{rabanser19}, the second comparing the distributions of the model predictions as proxies for the true prior distributions, called Black-Box Shift Detector (BBSD) \cite{lipton18}. The feature-based approaches are more sensitive to shifts impacting the feature distribution, while the BBSD has been specifically designed to deal with prior shift. In both cases, hypothesis testing is employed to seek statistically significant differences.
Although these two approaches have different designs and purposes, they are usually compared on generic shift detection tasks mainly on datasets for visual classification \cite{rabanser19}. 

The key idea of this paper is that ensembling shift detectors and adapting the significance level of their statistical tests is a more robust approach, especially suited for real-life settings where the precise shift type is unknown. This result is supported by the following contributions:
\begin{itemize}
\item We propose various ensemble schemes to combine feature- and prediction-based shift detectors and validate detectors ensembling as a more robust solution.
\item Motivated by the observation that the statistical tests needed for drift detection have a different behaviour across datasets, we propose the adaptation of the significance level required for detection to the specific dataset under study.
\item We perform a statistically sound benchmark study of base and ensembled shift detectors on structured datasets ($21$ OpenML\footnote{www.openml.org} and Kaggle \footnote{www.kaggle.com} classification datasets, largely differing in number of features and class proportions). Indeed the vast majority of deployed models in production consume structured data, highlighting the need for studies on model and data monitoring approaches for this modality. The shift detectors are evaluated against various types of synthetic drifts, including selection bias drift, which is less explored in the existing shift detection benchmarks although very frequent in real-world scenarios.
\item Last but not least, we also investigate both theoretically and empirically the degradation of the BBSD detector, a core component of the detectors ensembles, as the predictive power of the underlying classifier drops.
\end{itemize}

After presenting related work, Section \ref{sec:dd} describes the shift detection approaches in more details, as well as promoting the benefit of adapting the statistical tests to the dataset. Section \ref{sec:exp} presents the experimental setting with a focus on synthetic drift generation, while Section \ref{sec:res} discusses the main outcomes and observations from the experiments. 

\section{Related Work}
\label{sec:related}

Dataset shift \cite{Candela2009,Moreno-Torres2012} has been addressed extensively in the context of \textbf{domain adaptation}, by exploring techniques aiming to align the model representations of source and target data, especially without target labels \cite{Kouw2019,Redko2020}. 

Label shift correction is a sub-field of domain adaptation that recently contributed to novel shift detection approaches. In particular, \cite{lipton18} proposed a shift detection approach based on the Black Box Shift Learning technique to correct label shift. \cite{rabanser19} presented a general framework for shift detection and evaluated the BBSD and other state-of-the-art shift detection techniques on image datasets. A similar approach to correct prior shift is the Regularized Learning under Label Shifts \cite{Azizzadenesheli2019}, also relying on classifiers predictions. An alternative maximum likelihood approach to estimate target label distributions highlights the benefits of calibration under prior shift \cite{Alexandari2019,Garg2020}.  

Best practices for responsible deployment of ML models include \textbf{data monitoring}, by inspecting incoming data for any signs of deviation from the expected scenario \cite{Amodei2016,Klaise2020}. While dataset shift detection is an important part of ML monitoring system, recent work \cite{Schelter20,Elsahar19} has focused on directly predicting models performance drop for specific task-relevant drifts. 

Besides monitoring possible distributional changes, the analysis of individual samples against the source dataset introduces the \textbf{anomaly detection} problem, extensively treated in the machine learning literature \cite{Ruff2020,Chandola09,Markou03}. 

Another related field is \textbf{out-of-distribution detection} \cite{Shafaei2018}, which together with shift detection task has been used recently to evaluate ML models \textbf{predictive uncertainty} \cite{Ovadia2019}. The study of predictive uncertainty offers a unifying view of the robustness a ML model should be equipped with to address any type of unexpected abnormality in input data. Finally \textbf{hypothesis testing} \cite{Steinebach2006} is a core part of all shift detection techniques. 

\section{Shift Detectors}
\label{sec:dd}

\subsection{Notations and Problem Setup}
Let $\mathcal{X}$ be an input space and $\mathcal{Y}$ be a label space where $|\mathcal{Y}|$ is finite. Let $X\in \mathcal{X}$ and $Y\in \mathcal{Y}$ be random variables. We denote by $\mathit{f}: \mathit{X} \rightarrow \mathit{Y}$ a predictor of the classification task at hand and we will refer to it as the \textit{primary model}.

\subsection{Feature-Based Detection}

The simplest feature-based detection is univariate hypothesis testing on individual features. For $n$ features, this requires $n$ univariate hypotheses tests to separately compare the distributions of features between the source and the target domain, and eventually aggregate the $n$ $p$-values through Bonferroni correction.

However it can quickly become cumbersome when the feature space is large. For a more scalable solution, a more compact data representation can be obtained through standard dimensionality reduction techniques such as Principal Component Analysis (PCA) or Sparse Random Projection (SRP) \cite{rabanser19}. 

In this work we do not consider detectors based on multivariate hypothesis testing as they were found to offer comparable performance to aggregated univariate tests \cite{rabanser19}.

A different solution to detect feature distributional changes in high dimensional datasets relies on domain-discriminating models \cite{Ben-david2006}.
The \textit{domain classifier} is a model trained to discriminate source and target domains, using the very same features employed by the primary model. By design a domain classifier is meant to identify covariate shift and sample selection bias, but not prior shift which assumes unchanged feature distribution.

High accuracy of the classifier is a symptom of dataset shift, therefore a Binomial test with a null hypothesis of $0.5$ accuracy for indistinguishable domains and balanced datasets is used to make sure the observed difference is statistically significant. One limitation of the domain classifier is the need of retraining for any new incoming target dataset.

\subsection{Prediction-Based Detection}
\label{subsec:bbsd}

BBSD exploits the primary model to measure the distributions of predictions on source and target features, which are used as proxies for the true source and target label distributions. There are two variants of BBSD, one using the probability outputs (BBSDs), and the other using the hard-thresholded predictions (BBSDh).

For BBSDh, a $\chi^2$ test is used to compare source and target predicted class distributions, while for BBSDs the Kolmogorov-Smirnov (KS) test is employed to compare per-class probability distributions.
The primary model predictions are $k$-dimensional with $k= |\mathcal{Y}|$ the number of classes in the primary task, thus the BBSDs shift detection requires $k$ univariate hypothesis tests. As in the previous section, Bonferroni correction \cite{rabanser19} can be used to aggregate the $k$ KS tests. 

Complementarily to the domain classifier, the BBSD has been specifically designed to address prior shift situations, but the distribution of predictions can also serve as proxy for the distribution of covariates \cite{lipton18}.

\subsection{Limitations of Shift Detectors}
\label{subsec:adapt-sign-level}

\paragraph{BBSD degrades as primary performances drop.}
We present two limitations of the above shift detection method. Firstly, the statistical power of BBSD is limited by the predictive power of the primary model. We theoretically prove in the Appendix (cf. Section \ref{subsec:app-bbsd}) that there exist prior shifts such that the smaller the predictive power of the model is, the higher the $p$-values of the BBSD test are, reducing the sensitivity of the test. This phenomenon also occurs on more general type of shifts as validated experimentally (cf. Section \ref{sec:exp}).

\paragraph{Detection power reduced by Bonferroni correction.}
Another limitation stems from the use of Bonferroni correction. Indeed, when the $p$-values of $k$ hypotheses are aggregated by Bonferroni correction, a significance level of $\alpha/k$ is required for each hypothesis to ensure that the Type-I error rate stays below $\alpha$. However if the hypotheses are not independent, this correction is too conservative and the family-wise Type-I error is controlled at a level strictly lower than $\alpha$, resulting in a loss of power \cite{Nakagawa2004,Perneger1998}. This is the case for the BBSDs test as the underlying tests are done on the distribution of predicted probabilities of each class.  For binary classification, the null hypothesis for one class and  its complementary class are the same so that the Type-I error rate is actually bounded by $\alpha/2$ (instead of the expected $\alpha$), reducing the test's power.
This is also the case for the univariate hypothesis testing on individual features, where the Bonferroni correction is conservative because of features correlation.
We propose to address the latter limitation by adapting the significance level to the dataset.

\subsection{Dataset Adaptive Significance Level}
\label{subsec:adapt-sign-level}

To improve the BBSDs test's power, keeping the Type-I error below a desired rate (i.e. $5\%$), we set the significance level to the $5\%$ quantile of the empirical $p$-value distribution under the null hypothesis. We achieve this by performing $100$ runs of shift detection comparing different random splits of the source dataset only. 

A dataset-specific significance level is not only beneficial for shift detectors requiring aggregation of multiple univariate tests, but also when dealing with small datasets (i.e. from $10$ to $100$ samples).
Indeed, performing statistical tests on small datasets yields $p$-values that can only assume few values. This quantization leads to non-uniform distributions of $p$-values under the null hypothesis so that a dataset-specific significance level is also beneficial.

The drawback of this adaptation represents the initial setup cost of running detection experiments on the source data, but the selected significance level can be used for all subsequent shift detection tasks for the given dataset.

Finally, for very imbalanced datasets (\textit{creditcard}, \textit{pc2} and \textit{mc1} where the minority class represents less than $0.7\%$), the distribution of the BBSDs $p$-values under $H_0$ is extremely skewed towards $1$. The study of the impact of imbalanced datasets on prediction-based detectors is left for future research. 

As a natural extension of base shift detectors, the detectors ensembles also benefit from the adaptation of the significance level.

\subsection{Detectors Ensembles}
\label{subsec:ensembles}

We evaluate two ensembling strategies combining feature and prediction-based shift detectors, in order to exploit their complementary detection abilities:
\begin{itemize}
\item domain classifier trained on both features and primary model predictions;
\item pair of statistical tests from complementary shift detectors with Bonferroni correction.
\end{itemize}

Ensembling strategies are naturally robust to various shift types, and represent a better fit to real-life settings where the precise shift type is often unknown. We confirm experimentally that ensembling retains the advantages of feature- and prediction-based shift detectors (cf. Section \ref{sec:exp}), providing a single practical tool for reliable drift monitoring. 

\section{Experimental Setup}
\label{sec:exp}

The purpose of our experiments is to compare the different detection approaches on various types of simulated drift over a large collection of structured datasets. Specifically we aim at highlighting differences in the approaches related to both the detection power and efficiency, the latter expressed in terms of minimum number of samples required to detect a shift. All the details of the experiments setup and reproducibility of this work are available in the Appendix (cf. Section \ref{subsec:app-exp}).

\subsection{Datasets and Shift Simulation}

In our experiments we use $21$ datasets for classification tasks from OpenML and Kaggle as listed in Table \ref{tab:datasets} of the Appendix. 

In order to simulate dataset drift we synthetically apply different types of shift to a split of the original dataset. The $10$ simulated shifts belong to the following categories and are detailed in Table \ref{tab:drift-types}: 
\begin{itemize}
\item Prior shift: generated by changing the fraction of samples belonging to a class.
\item Covariate shift due to Gaussian noise: generated by adding Gaussian noise to some numeric features of a fraction of samples.
\item Covariate shift due to Adversarial noise: generated by applying Adversarial noise to numeric features. It's a more subtle kind of noise, slightly changing the features but inducing the primary model to switch its predicted class.
\item Selection bias: generated by selecting samples with a probability dependent on the sample features and implicitly dependent on the class, possibly over-sampling by interpolation of existing observations.
\end{itemize}

The severity of those shifts is controlled by the parameters shown in Table \ref{tab:drift-types}. Our experiments explore the effect of changing those parameters values generating a total of $19$ drifts, then averaging the outcome from the same shift type to build the results for the $10$ shifts types in Table \ref{tab:drift-types}.

\begin{table}[t]
\caption{Simulated shift types.}
\label{tab:drift-types}
\begin{adjustbox}{width=\columnwidth,center}
\begin{tabular}{lllll}
\toprule
Category & Shift type &  \multicolumn{2}{c}{Parameters} & Description \\
\midrule
\multirow{3}{*}{Prior shift} & \multirow{2}{*}{Knock-out} & $s=25\%$ & & Remove $s$\% of majority class. \\
    &  & $s=40\%$ & & \\
\cmidrule{2-4}
    & Only-One &  & & Select only-one minority class. \\
\midrule
\multirow{12}{*}{Covariate shift} & \multirow{4}{*}{Small Gaussian}  & $s=50\%$ & $f=50\%$ & Small amount of Gaussian Noise applied on $s$\% of samples \\
& & $s=50\%$ & $f=100\%$ &  and $f$\% of features. \\
& & $s=100\%$ & $f=50\%$ &  \\
& & $s=100\%$ & $f=100\%$ &  \\
\cmidrule{2-4}
 & \multirow{4}{*}{Medium Gaussian}  & $s=50\%$ & $f=50\%$ & Medium amount of Gaussian Noise applied on $s$\% of samples \\
 &  & $s=50\%$ & $f=100\%$ & and $f$\% of features. \\
 & & $s=100\%$ & $f=50\%$ &  \\
& & $s=100\%$ & $f=100\%$ &  \\
\cmidrule{2-4}
    & \multirow{2}{*}{Adversarial ZOO} & $s=25\%$ & & Zeroth-order optimization black box adversarial attack\\
    & & $s=50\%$ & & on $s$\% of samples \cite{Chen2017} .\\
    \cmidrule{2-4}
    & \multirow{2}{*}{Adversarial Boundary} & $s=25\%$ & & Boundary black box adversarial attack \\
    & & $s=50\%$ & & on $s$\% of samples \cite{Brendel2018}.\\
\midrule
\multirow{8}{*}{Selection bias} & \multirow{2}{*}{Joint Subsampling} & &  & Keeps an observation with probability decreasing as points  \\
 & & & & are  away from the samples mean. \\
 \cmidrule{2-4}
    & \multirow{2}{*}{Subsampling} & & $f=100\%$ & Subsample with low probability samples with low feature \\
     & &&  & values separately for $f$\%  features.  \\
     \cmidrule{2-4}
    & \multirow{2}{*}{Under-sampling} & $s=50\%$ & & Keep $s$\% of samples, selecting samples close to the \\
     & & & & minority class (NearMiss3 heuristics).  \\
     \cmidrule{2-4}
    & \multirow{2}{*}{Over-sampling} & $s=50\%$ & & Replace $s$\% of samples with samples interpolated from \\
    & &&  & the remaining part.  \\
\bottomrule
\end{tabular}
\end{adjustbox}
\end{table}

The combination of different shift types is also relevant as it can occur in real-life settings, for this reason the study of the detectors response to composite shifts will be considered in future work. 

\subsection{Experiments}

For each run of drift experiment, we test all the mentioned shift detection approaches on subsets of the source and target datasets with number of samples in $[10, 100, 500, 1000, 2000]$. 
The different detection approaches are reported using the following labels:
\begin{itemize}
\item \textit{BBSDs}: soft version of BBSD with Random Forest primary model and KS test.
\item \textit{BBSDh}: hard version of BBSD with Random Forest primary model and $\chi^2$ test.
\item \textit{Test\_X}: KS test on input features. 
\item \textit{Test\_PCA}: KS test on PCA-projected features.
\item \textit{Test\_SRP}: KS test on SRP-projected features.
\item \textit{DC}: Random Forest domain classifier with Binomial test. 
\end{itemize}

The ensembling strategies are reported using the following labels:
\begin{itemize}
\item \textit{BBSDs + X}: \textit{BBSDs} and \textit{Test\_X} with Bonferroni correction.
\item \textit{BBSDs + DC}: \textit{BBSDs} and \textit{DC} with Bonferroni correction.
\item \textit{DC*}: \textit{DC} trained on both features and primary predictions.
\end{itemize}

For any detector, its adaptive variant using a significance level tailored to the dataset (cf. Subsection \ref{subsec:adapt-sign-level}) is referred to with the suffix \textit{(adapt)}, i.e. \textit{BBSDs (adapt)}.

Overall, we have collected $p$-values from shift detection experiments for $19$ different drifts, run $5$ times, for $5$ different sizes, for each of the $21$ datasets. Furthermore the results over the different runs are averaged, as well as the results from the same types of drift with different drift parameters to yield $210$ ($21$ datasets $\times 10$ shift types) detection results per dataset size per detector. 

\subsection{Metrics}
\label{subsec:metrics}

The $p$-values of the detectors are averaged over multiple runs and when their value is less than the significance level, we consider that a drift has been detected. The significance level used in the non-adaptive version of the statistical test is $0.05$.

Considering shift detectors as binary classifiers where one sample is one comparison of source and target datasets, we can measure its accuracy and true positive rate (TPR). When the accuracy is reported the evaluated shift scenarios include a negative (no shift) situation per dataset.

To measure the data efficiency of a shift detector, we define an \textbf{efficiency score} as the minimum size level required to detect the shift. As the detectors are evaluated at different datasets sizes $[10, 100, 500, 1000, 2000]$, they are assigned a score from $5$ to $1$. A detector failing to find the shift at any size is assigned a score of $0$.  

\subsection{Statistical Comparison}

In order to fairly compare the detection approaches, we use a Friedman test to statistically assess whether the detectors have different performances. When the conclusion is positive, a Nemenyi post-hoc test is used to determine pairwise equivalence of the methods \cite{Demsar2006}. 

Based on the efficiency score defined in Section \ref{subsec:metrics}, the average ranks for each detector are computed across all datasets. The Friedman test then checks whether the detectors average ranks are significantly different from the mean rank expected under the null-hypothesis. When the former test rejects the null-hypothesis, the Nemenyi test is used to seek for pairwise differences between ranks.

In order to have independent observations from different datasets, we perform these statistical tests separately for each of the $10$ shift types. We use the library \textit{autorank}\footnote{\small \url{https://github.com/sherbold/autorank}} to perform this statistical comparison.

\section{Results}
\label{sec:res}
We first give the experiment results comparing ensembled and base shift detectors over all datasets and by shift types. We then highlight the complementarity of feature- and prediction-based shift detectors motivating the ensembling strategy. Finally, we present ablation studies comparing base detectors and ensembles to their adaptive counterparts.

\subsection{Ensembling Shift Detectors}
\label{subsec:res_ensembles}

\begin{figure}[t]
\begin{center}
\begin{minipage}[t]{.45\linewidth}
  \centering
\centerline{\includegraphics[width=\textwidth]{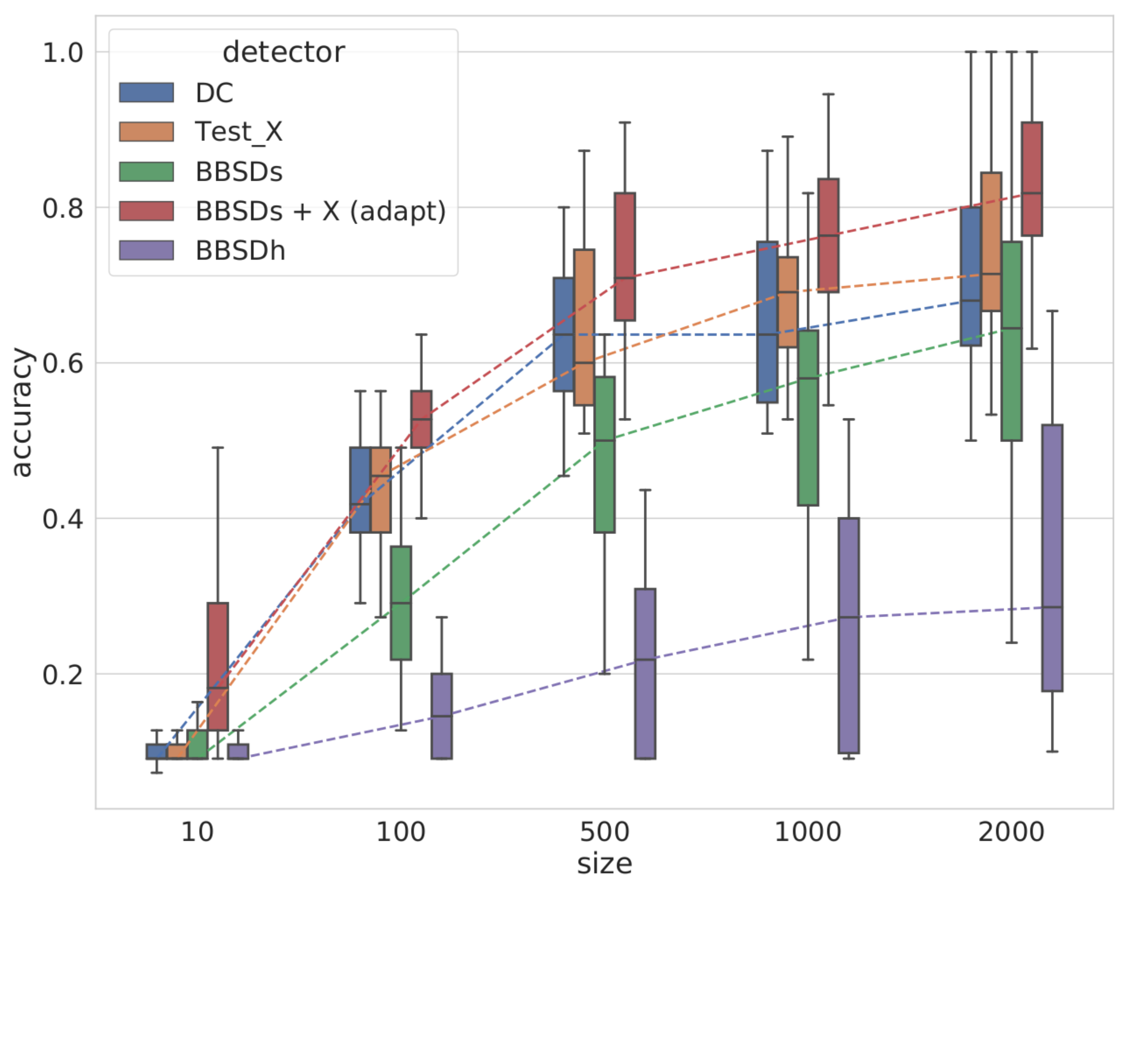}}
\caption{Shift detectors accuracy by dataset size.}
\label{fig:accuracy-boxplots}
\end{minipage}%
\hfill
\begin{minipage}[t]{.45\linewidth}
  \centering
\centerline{\includegraphics[width=\textwidth]{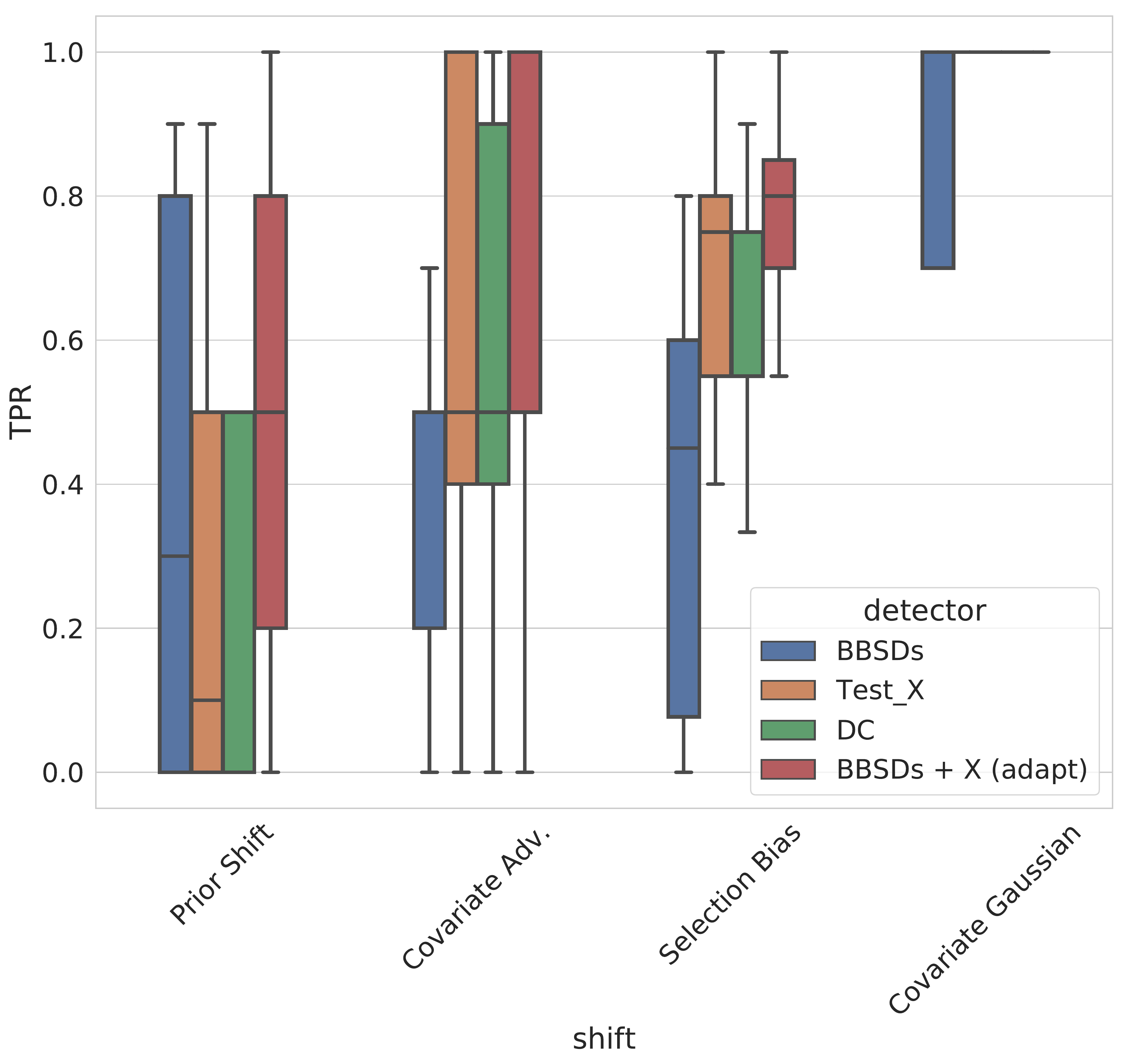}}
\caption{Shift detectors true positive rate (TPR) by type of shift.}
\label{fig:tpr-boxplots}
\end{minipage}
\end{center}
\end{figure}

When observing the average detection accuracy across datasets and shift types for different dataset sizes (Figure  \ref{fig:accuracy-boxplots}) some simple detection approaches stand out (\textit{Test\_X} and \textit{DC}), although slightly less accurate than the adaptive ensemble technique (\textit{BBSDs+X (adapt)}). However the true positive rate for different shift types for a fixed size of $1000$ (Figure \ref{fig:tpr-boxplots}) reveals a more complex landscape, where the performance of base shift detectors is not uniform. For an easier reading of Figures \ref{fig:accuracy-boxplots} and \ref{fig:tpr-boxplots}, some detectors have been omitted and the various shifts have been grouped by type (cf. Table \ref{tab:drift-types}), but all the detailed TPR and accuracy results for individual shift detectors and shift types can be found in Tables \ref{tab:acc-size} and \ref{tab:tpr-shift} of the Appendix.

In Tables \ref{tab:acc-size-ens} and \ref{tab:tpr-shift-ens} we report the mean accuracy by dataset size and mean TPR by type of shift for the base detectors and the detectors ensembles with adaptation of the significance level to the dataset. The ensemble \textit{BBSDs+X (adapt)} comes out as the most accurate shift detector overall.

In addition to be sensitive at lower sizes \textit{BBSDs+X (adapt)} is also able to capture alone most types of shifts. This approach represents a more robust solution able to detect the shift situations missed by either of the base shift detectors. 
It is interesting to notice that simpler ensembles \textit{BBSDs+X (adapt)} and \textit{BBSDs+DC (adapt)} perform very similarly to \textit{DC*}, while not requiring an additional training.  Ensembling feature- and prediction-based shift detectors is a promising strategy for an effective and robust drift monitoring.  

\begin{table}[h!]
\caption{Accuracy across all datasets by size (mean $\pm$ std). Best values and all values within $95$\% confidence interval are in bold.}
\label{tab:acc-size-ens}
\begin{adjustbox}{width=0.7\columnwidth,center}
\begin{tabular}{lcccccc}
\toprule
          &  BBSDs & Test\_X & DC &  \multicolumn{3}{c}{Ensembles (adapt)}  \\
Size &   &   &     &  DC* & BBSDs+DC & BBSDs+X  \\
\midrule
   10 & 0.11 \tiny{$\pm$ 0.03} & 0.10 \tiny{$\pm$ 0.02} & 0.10 \tiny{$\pm$ 0.02} &  \textbf{0.20} \tiny{$\pm$ 0.24} & 0.14 \tiny{$\pm$ 0.03} &  \textbf{0.22} \tiny{$\pm$ 0.11} \\
  100 & 0.30 \tiny{$\pm$ 0.10} & 0.45 \tiny{$\pm$ 0.10} & 0.44 \tiny{$\pm$ 0.09} & 0.46 \tiny{$\pm$ 0.17} &  \textbf{0.50} \tiny{$\pm$ 0.10} &  \textbf{0.54} \tiny{$\pm$ 0.10} \\
  500 & 0.46 \tiny{$\pm$ 0.14} & 0.65 \tiny{$\pm$ 0.12} & 0.64 \tiny{$\pm$ 0.10} & 0.63 \tiny{$\pm$ 0.14} & 0.67 \tiny{$\pm$ 0.12} &  \textbf{0.73} \tiny{$\pm$ 0.11} \\
 1000 & 0.54 \tiny{$\pm$ 0.16} & 0.69 \tiny{$\pm$ 0.11} & 0.66 \tiny{$\pm$ 0.11} & 0.63 \tiny{$\pm$ 0.13} &  \textbf{0.72} \tiny{$\pm$ 0.12} &  \textbf{0.75} \tiny{$\pm$ 0.12} \\
 2000 & 0.63 \tiny{$\pm$ 0.19} & 0.75 \tiny{$\pm$ 0.13} & 0.71 \tiny{$\pm$ 0.14} & 0.67 \tiny{$\pm$ 0.15} &  \textbf{0.79} \tiny{$\pm$ 0.12} &  \textbf{0.83} \tiny{$\pm$ 0.12} \\
\bottomrule
\end{tabular}
\end{adjustbox}
\end{table}

\begin{table}[h!]
\caption{TPR across all datasets by shift type at dataset size of $1000$ (mean $\pm$ std). Best values and all values within $95$\% confidence interval are in bold.}
\label{tab:tpr-shift-ens}
\begin{adjustbox}{width=0.8\columnwidth,center}
\begin{tabular}{lcccccc}
\toprule
          &  BBSDs & Test\_X & DC &  \multicolumn{3}{c}{Ensembles (adapt)}  \\
Shift type &   &   &     & DC* &  BBSDs+DC & BBSDs+X  \\
\midrule
 Knock-Out         & \textbf{0.23} \tiny{$\pm$ 0.30} & 0.11 \tiny{$\pm$ 0.24} & 0.00 \tiny{$\pm$ 0.00} & 0.01 \tiny{$\pm$ 0.05} & \textbf{0.25} \tiny{$\pm$ 0.30} & \textbf{0.34} \tiny{$\pm$ 0.40} \\
 Only-One          & \textbf{0.60} \tiny{$\pm$ 0.46} & 0.55 \tiny{$\pm$ 0.49} & 0.46 \tiny{$\pm$ 0.47} & 0.51 \tiny{$\pm$ 0.48} & \textbf{0.70} \tiny{$\pm$ 0.44} & \textbf{0.73} \tiny{$\pm$ 0.41} \\
 Small Gaussian    & 0.74 \tiny{$\pm$ 0.37} & 0.96 \tiny{$\pm$ 0.14} & \textbf{1.00} \tiny{$\pm$ 0.00} & 0.95 \tiny{$\pm$ 0.12} & \textbf{1.00} \tiny{$\pm$ 0.00} & 0.99 \tiny{$\pm$ 0.04} \\
 Medium Gaussian   & 0.87 \tiny{$\pm$ 0.32} & 0.98 \tiny{$\pm$ 0.06} & \textbf{1.00} \tiny{$\pm$ 0.00} & \textbf{1.00} \tiny{$\pm$ 0.00} & 0.99 \tiny{$\pm$ 0.04} & 0.99 \tiny{$\pm$ 0.04} \\
 Adv. ZOO          & 0.21 \tiny{$\pm$ 0.40} & \textbf{0.43} \tiny{$\pm$ 0.51} & \textbf{0.42} \tiny{$\pm$ 0.50} & \textbf{0.47} \tiny{$\pm$ 0.50} & \textbf{0.43} \tiny{$\pm$ 0.51} & \textbf{0.45} \tiny{$\pm$ 0.49} \\
 Adv. Boundary     & 0.72 \tiny{$\pm$ 0.42} & 0.77 \tiny{$\pm$ 0.35} & 0.70 \tiny{$\pm$ 0.41} & 0.72 \tiny{$\pm$ 0.40} & 0.74 \tiny{$\pm$ 0.40} & \textbf{0.98} \tiny{$\pm$ 0.06} \\
 Subsampling Joint & 0.04 \tiny{$\pm$ 0.08} & 0.20 \tiny{$\pm$ 0.30} & 0.10 \tiny{$\pm$ 0.19} & \textbf{0.30} \tiny{$\pm$ 0.35} & 0.16 \tiny{$\pm$ 0.26} & \textbf{0.34} \tiny{$\pm$ 0.30} \\
 Subsampling       & 0.42 \tiny{$\pm$ 0.41} & \textbf{0.71} \tiny{$\pm$ 0.45} & \textbf{0.70} \tiny{$\pm$ 0.44} & 0.17 \tiny{$\pm$ 0.29} & \textbf{0.72} \tiny{$\pm$ 0.43} & \textbf{0.72} \tiny{$\pm$ 0.45} \\
 Under-sampling    & 0.44 \tiny{$\pm$ 0.44} & \textbf{1.00} \tiny{$\pm$ 0.00} &\textbf{1.00} \tiny{$\pm$ 0.00} & \textbf{1.00} \tiny{$\pm$ 0.00} & \textbf{1.00} \tiny{$\pm$ 0.00} & \textbf{1.00} \tiny{$\pm$ 0.00} \\
 Over-sampling     & 0.57 \tiny{$\pm$ 0.40} & \textbf{1.00} \tiny{$\pm$ 0.00} & \textbf{1.00} \tiny{$\pm$ 0.00} & \textbf{1.00} \tiny{$\pm$ 0.00} & \textbf{1.00} \tiny{$\pm$ 0.00} & \textbf{1.00} \tiny{$\pm$ 0.00} \\
\bottomrule
\end{tabular}
\end{adjustbox}
\end{table}

\subsection{Comparison of Base Shift Detectors}
\label{subsec:res_bench}

In order to have a global comparison of the performance of base detectors, we aggregate the results from different dataset sizes with the efficiency score and show the global average ranks by type of shift in Table \ref{tab:avg-ranks}. A first observation is that overall the feature-based detectors have lower rank, indicating that they are more effective than other approaches, requiring less data to detect a drift situation.

\begin{figure}[h]
\begin{center}
\centerline{\includegraphics[width=0.8\linewidth]{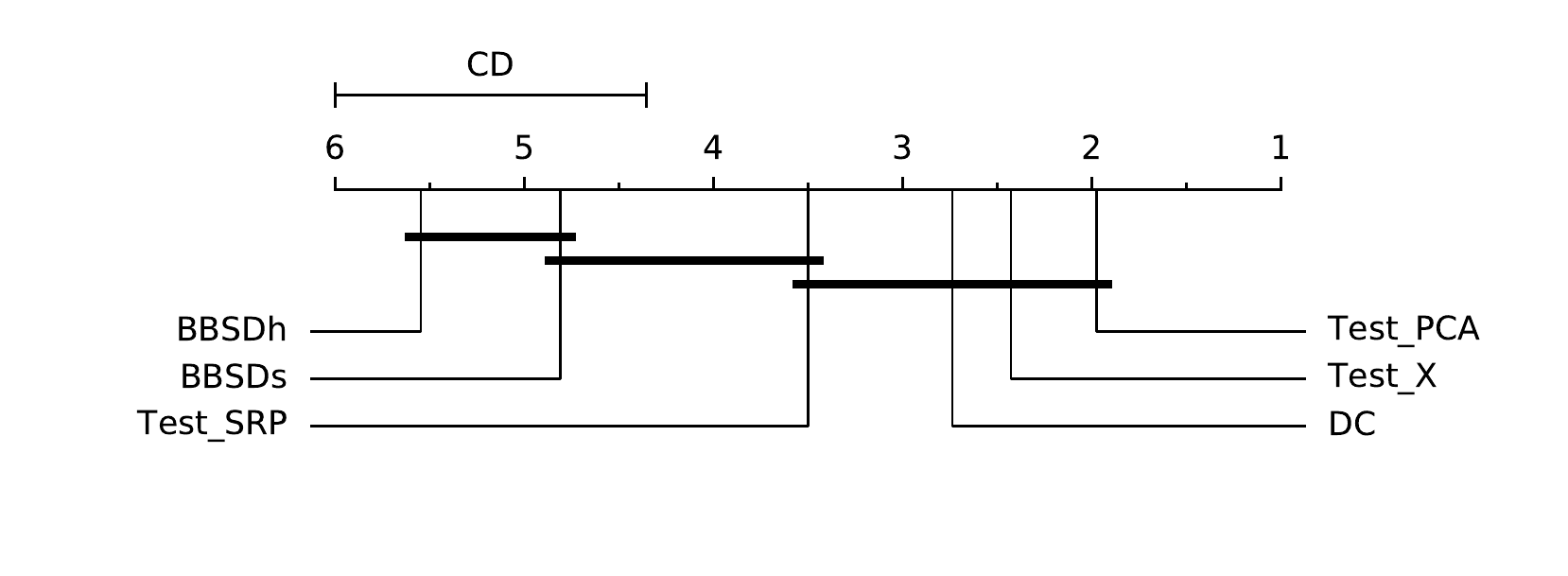}  }
\caption{Pairwise comparison of all detectors with the Nemenyi test for \textit{Under-sampling} shift.}
\label{fig:nemenyi_ex}
\end{center}
\end{figure}

\begin{table*}[t]
\caption{Average ranks of $6$ shift detectors based on efficiency score.}
\label{tab:avg-ranks}
\begin{adjustbox}{width=0.7\columnwidth,center}
\begin{tabular}{lcccccc}
\toprule
Shift type & BBSDs & BBSDh & Test\_X &  Test\_PCA & Test\_SRP  & DC\\
\midrule
Knock-Out & \textbf{2.71} & 3.02 & 3.74 & 3.67 & 3.62 & 4.24 \\
Only-One  & \textbf{2.64} & 3.07 & 3.45 & 3.90 & 3.90 & 4.02 \\
Small Gaussian   & 2.93 & 5.43 & 1.95 & 4.93 & 4.26 & \textbf{1.50} \\
Medium Gaussian   & 2.36 & 5.52 & 2.26 & 4.79 & 4.14 & \textbf{1.93} \\
Adversarial ZOO  & 3.81 & 4.21 & \textbf{2.60} & 3.93 & 3.40 & 3.05 \\
Adversarial Boundary & 2.95 & 4.55 & \textbf{2.31} & 4.64 & 3.55 & 3.00 \\
Joint Subsampling & 3.98 & 3.98 & 3.26 & \textbf{2.52} & 3.29 & 3.98 \\
Subsampling  & 4.12 & 4.86 & \textbf{2.69} & 3.10 & 3.00 & 3.24 \\
Under-sampling  & 4.81 & 5.55 & 2.43 & \textbf{1.98} & 3.50 & 2.74 \\
Over-sampling  & 4.90 & 5.69 & 2.26 & \textbf{2.14} & 3.52 & 2.48 \\
\midrule
All &  3.54 & 4.63 &   \textbf{2.63} & 3.58 & 3.65 & 2.96 \\
\bottomrule
\end{tabular}
\end{adjustbox}
\end{table*}
 
Indeed for all the types of shift the Friedman test at $p=0.05$ reported that the approaches are not statistically equivalent. 

We thus look at the results of the post-hoc Nemenyi test, represented  in Figure \ref{fig:nemenyi_ex} for the case of under-sampling shift. The Critical Distance (CD) between the average ranks reveals two groups of different detectors, with feature-based detectors outperforming the BBSD approaches. The \textit{Test\_PCA}, \textit{Test\_X}, \textit{DC} detectors have statistically higher efficiency than \textit{BBSDh} and \textit{BBSDs}. The Nemenyi test at $p=0.05$ is not powerful enough to draw conclusions about the other detectors.

\begin{figure*}[h!]
\centering
\subfloat[Knock-Out shift]{
  \includegraphics[width=.44\linewidth]{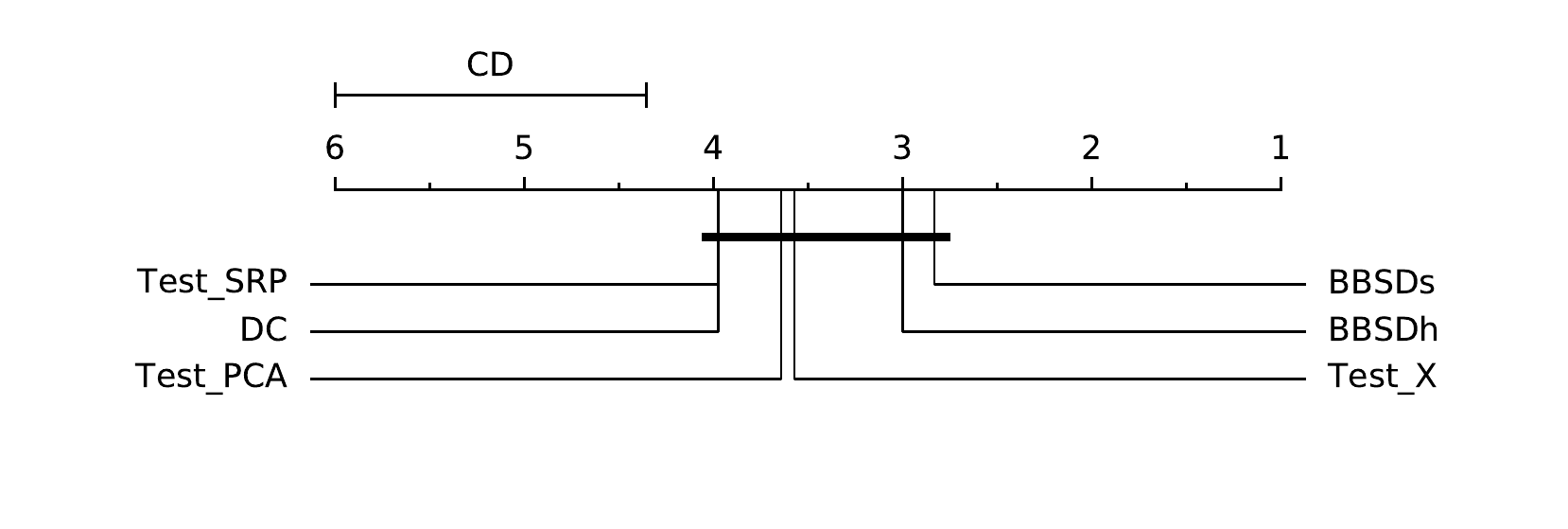}  
  \label{fig:ko}
}
\subfloat[Only-One shift]{
  \includegraphics[width=.44\linewidth]{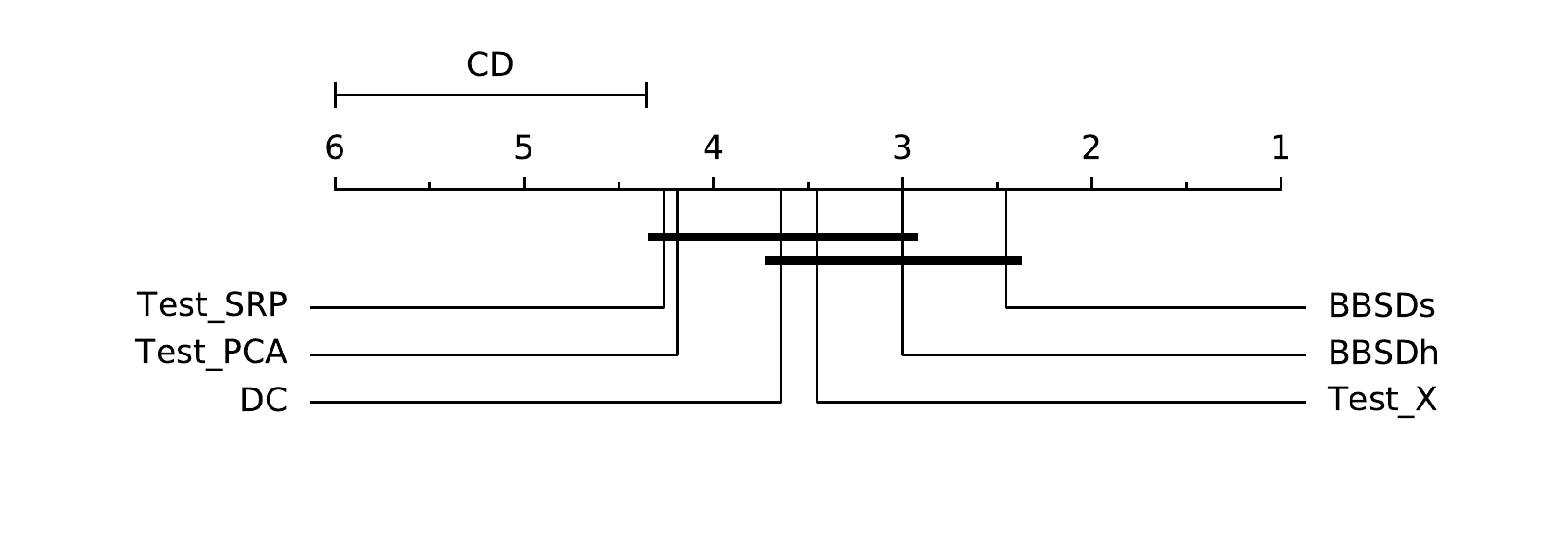}  
  \label{fig:oo}
}\\
\subfloat[Small Gaussian Noise shift]{
  \includegraphics[width=.44\linewidth]{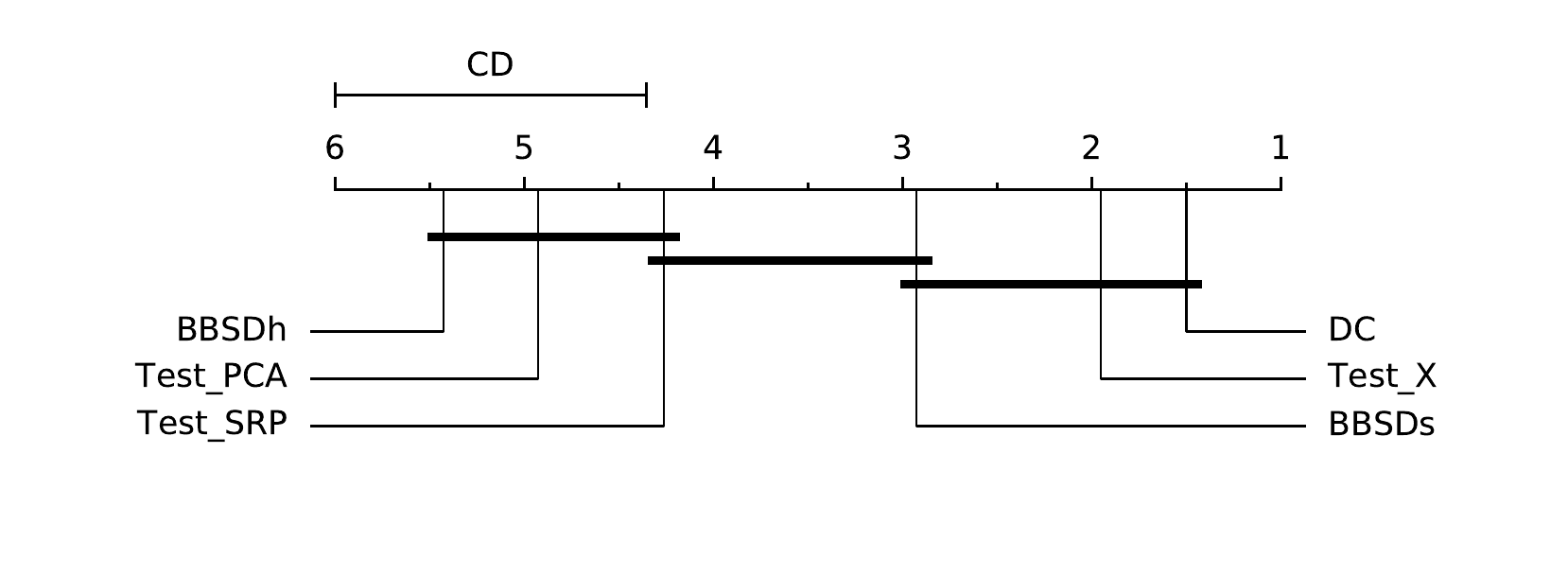}  
  \label{fig:sg}
}
\subfloat[Medium Gaussian Noise shift]{
  \includegraphics[width=.44\linewidth]{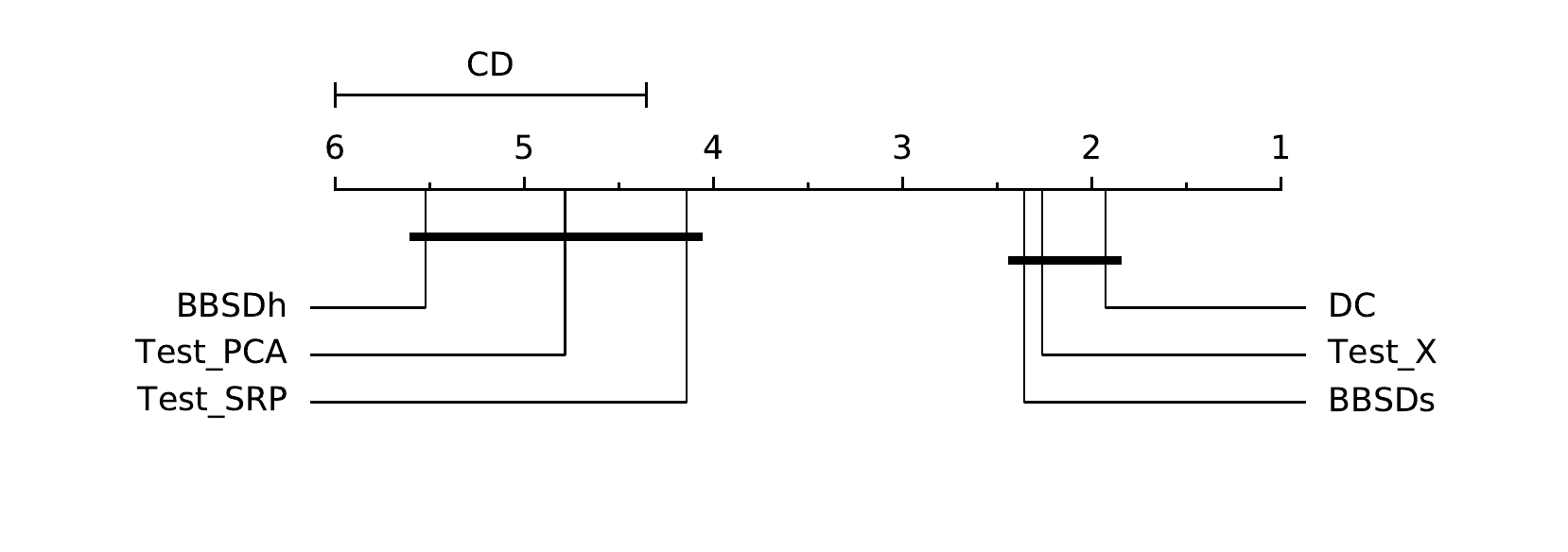}  
  \label{fig:mg}
}\\
\subfloat[Adversarial ZOO perturbation]{
  \includegraphics[width=.44\linewidth]{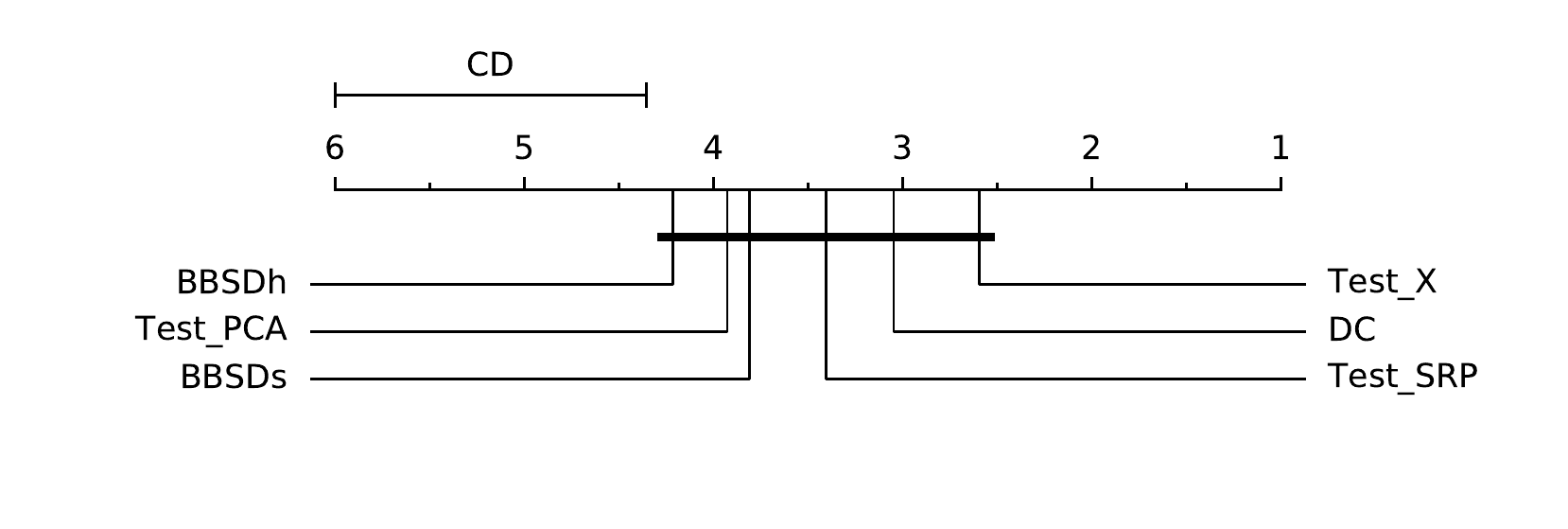}  
  \label{fig:advz}
}
\subfloat[Adversarial Boundary perturbation]{
  \includegraphics[width=.44\linewidth]{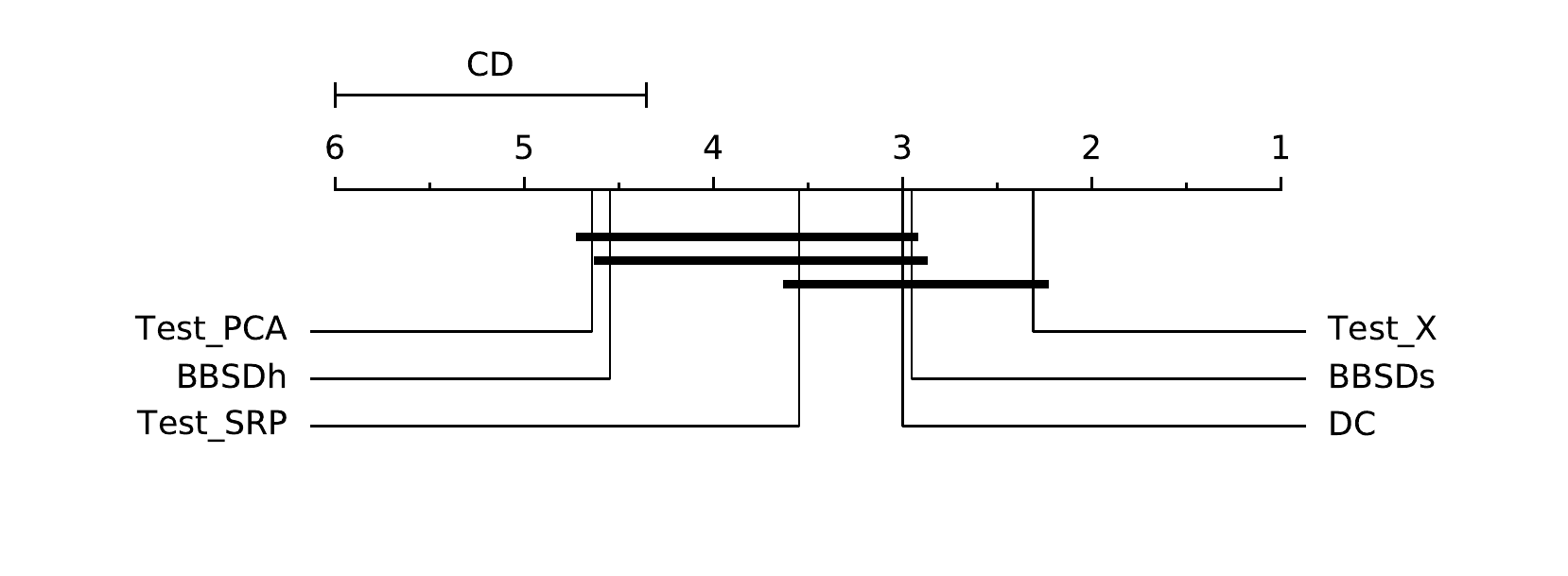}  
  \label{fig:advb}
}\\
\subfloat[Joint Subsampling shift]{
  \includegraphics[width=.44\linewidth]{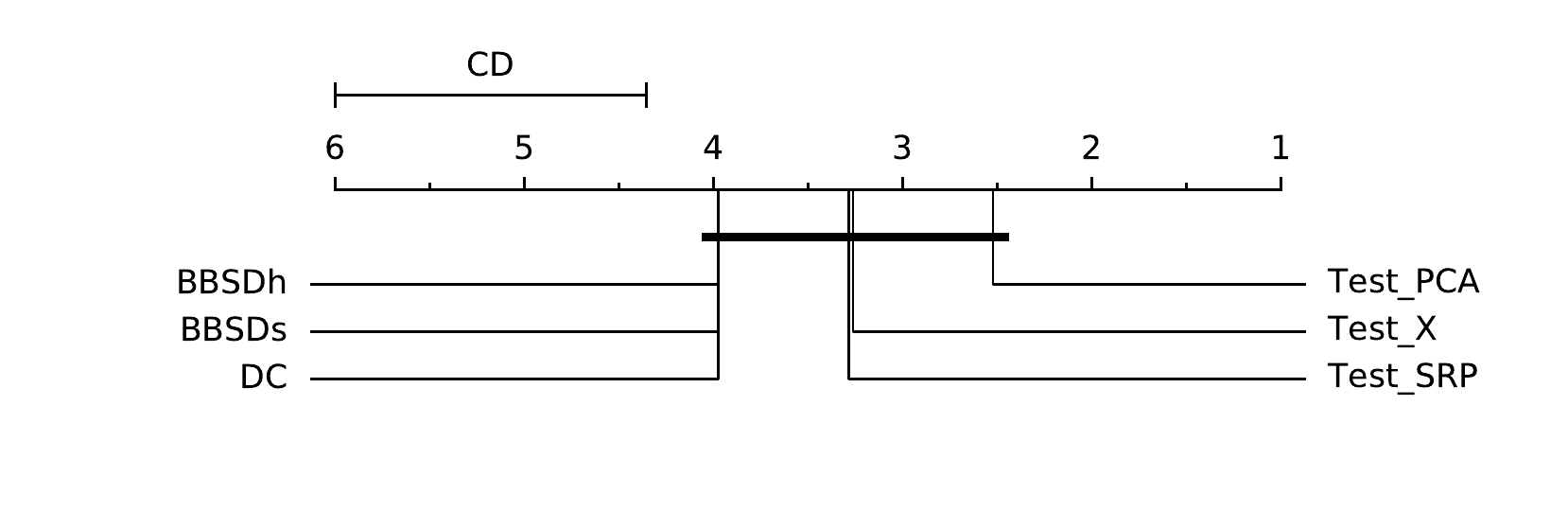}  
  \label{fig:js}
}
\subfloat[Subsampling shift]{
  \includegraphics[width=.44\linewidth]{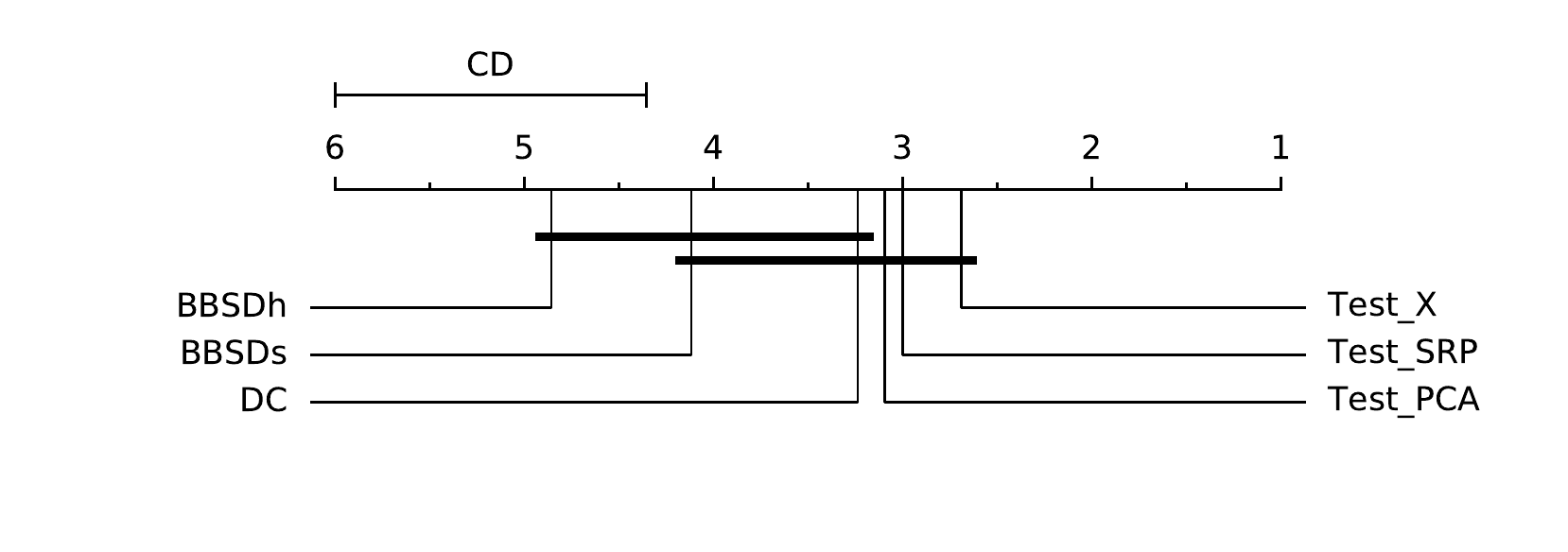}  
  \label{fig:ss}
}\\
\subfloat[Under-sampling shift]{
  \includegraphics[width=.44\linewidth]{figures/under_sample_shift}  
  \label{fig:us}
}
\subfloat[Over-sampling shift]{
  \includegraphics[width=.44\linewidth]{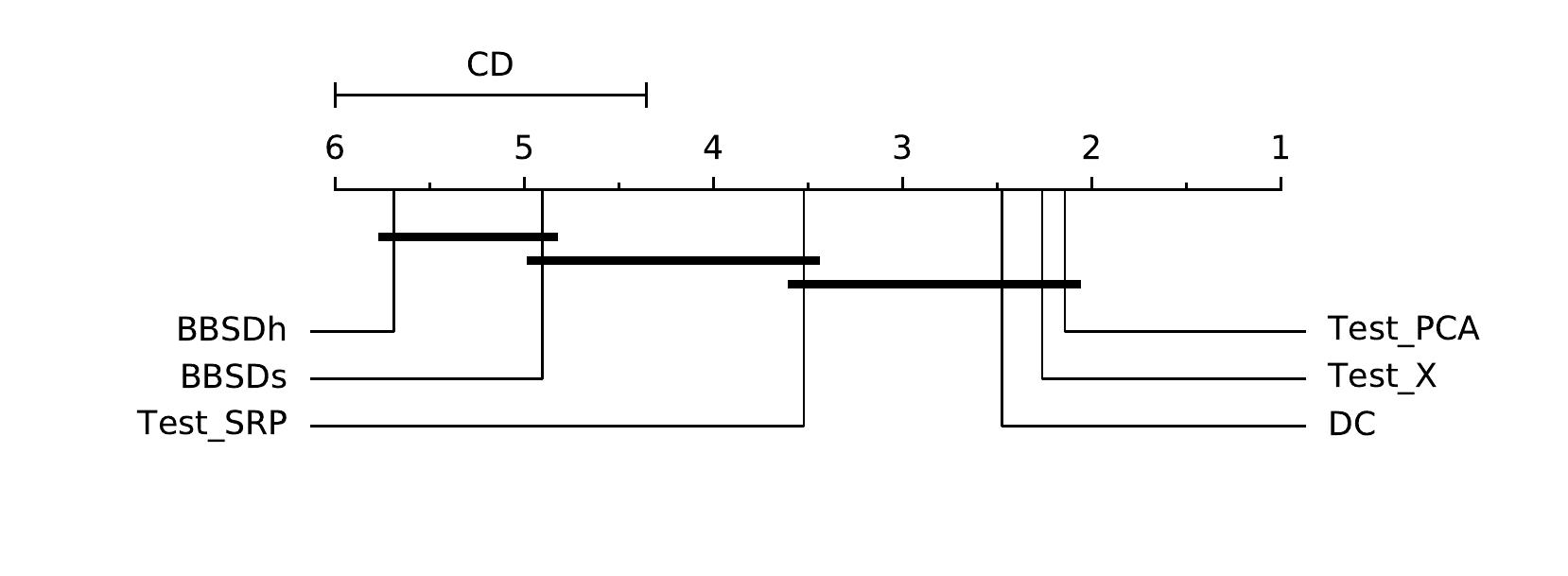}  
  \label{fig:os}
}
\caption{Nemenyi post-hoc tests for the different types of drift.}
\label{fig:nemenyi-plots}
\vskip -0.2in
\end{figure*}

Overall the Nemenyi tests (Figure \ref{fig:nemenyi-plots}) highlight the following differences:
\begin{itemize}
\item \textit{Prior shift}: prediction-based detectors perform better than other approaches when the label distribution is affected, as we could expect by design.
\item \textit{Gaussian noise}: domain classifier and direct input features testings are more effective than other approaches.
\item \textit{Adversarial Boundary}: BBSDs and direct feature testing perform slightly better than domain classifier and significantly better than BBSDh. This perturbation is too subtle to be spotted by the domain classifier, although its important impact on class distributions is easily detected by both BBSDs and Test\_X.
\item \textit{Subsampling, Under-sampling, Over-sampling}: features-based detectors perform significantly better than prediction-based detectors on perturbations simulating selection bias. 
\end{itemize}

The previous findings highlight the complementarity of feature- and prediction-based drift detectors on different drift scenarios, motivating the proposed ensembling strategy (cf. Section \ref{subsec:ensembles}).

The individual results from each drift scenario per dataset are available in the paper repository \footnote{\url{https://github.com/dataiku-research/drift_detectors_benchmark/disaggregated_results.md}} for full inspection.

\subsection{Impact of Dataset-Adaptive Significance Level}
\label{subsec:impact_adap}
\begin{figure}[t]
\begin{center}
\begin{minipage}[t]{.45\linewidth}
  \centering
\centerline{\includegraphics[width=\textwidth]{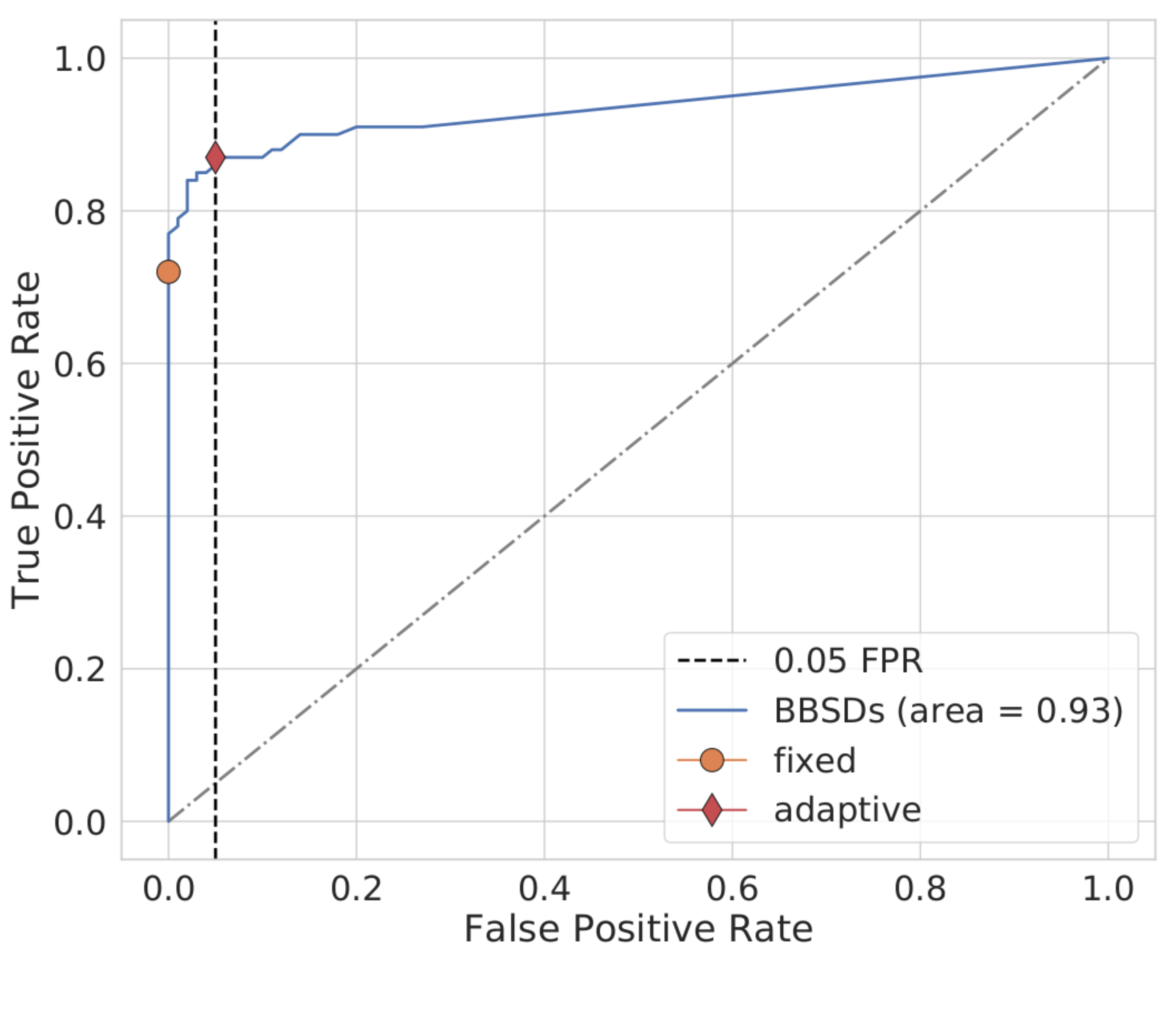}}
\caption{ROC curve of BBSDs as a binary classifier on all positive and negative drift experiments for the \textit{MagicTelescope} dataset.}
\label{fig:roc-adapt}
\end{minipage}%
\hfill
\begin{minipage}[t]{.45\linewidth}
  \centering
\centerline{\includegraphics[width=\textwidth]{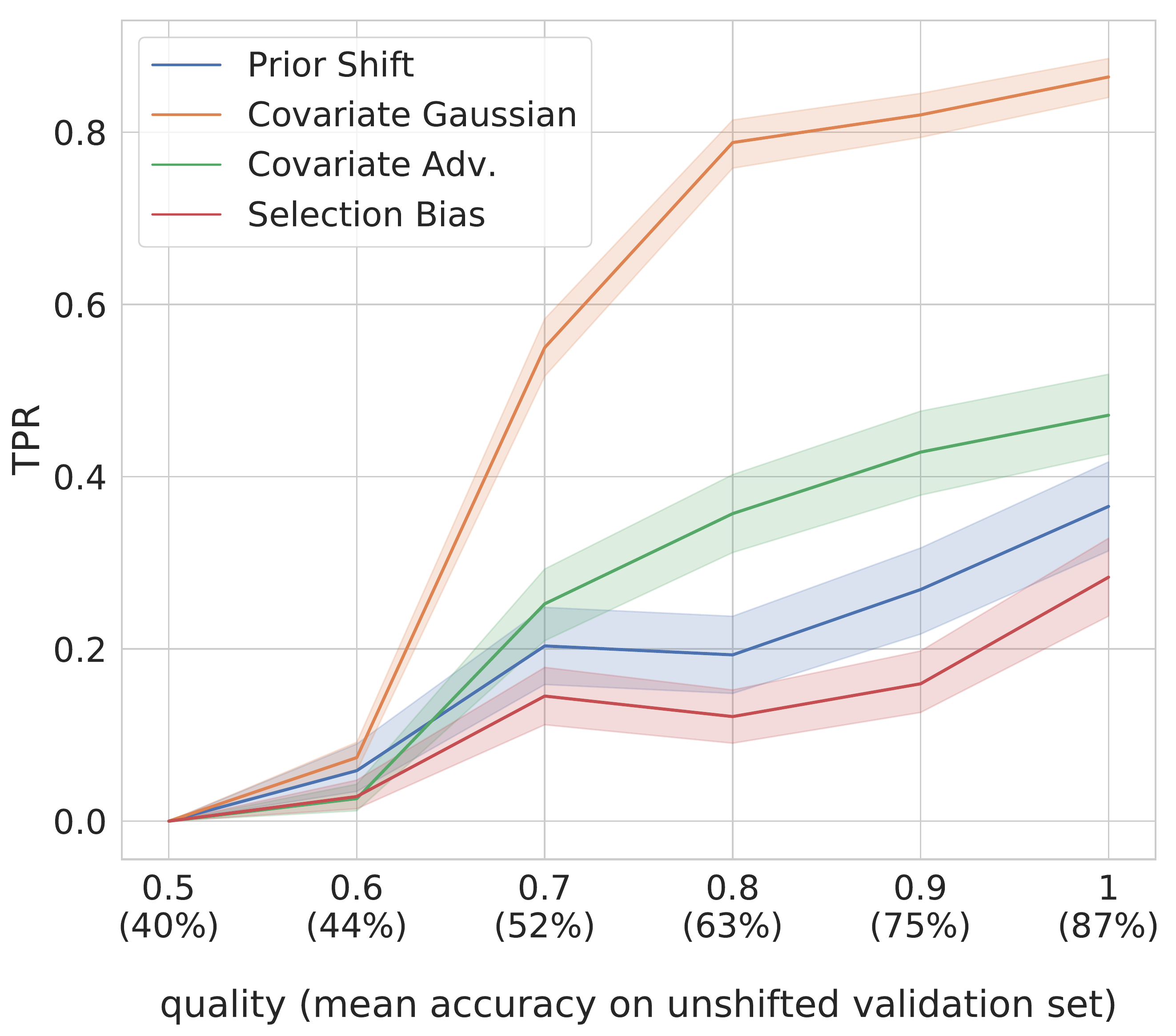}}
\caption{Average TPR with standard deviation from the BBSDs test with perturbed primary model. Higher corruption probability degrades the model performances.}
\label{fig:quality-boxplot}
\end{minipage}
\end{center}
\end{figure}

The dataset-adaptive significance level in BBSDs is consistently higher than the Bonferroni-corrected significance level of $\alpha/k$. For instance, for all binary classification datasets in Table \ref{tab:datasets} of the Appendix, the standard Bonferroni-corrected significance level is $\alpha/2=0.025$, while on average the computed adaptive level is $0.089$: this gap shows the margin to improve the test's power,  keeping the Type-I error below $5\%$. This is illustrated in Figure \ref{fig:roc-adapt} on the \textit{MagicTelescope} dataset, showing the BBSD detector as a binary classifier on all (positive) shift detection tests at size $2000$ described in the experiments and additional $100$ (negative) shift detection tests on randomly sampled validation and test sets of the same size, on which no shift is applied. 

In order to analyze the improvement achieved by adapting the significance level of the statistical tests as described in Section \ref{subsec:adapt-sign-level}, we compare the regular and adaptive versions of the top-3 base shift detectors through Friedman and Nemenyi post-hoc test. Table \ref{tab:adapt-avg-ranks} and Figure \ref{fig:adapt-nemenyi-plots} (in the Appendix) highlight the improved sensitivity of the adaptive detectors. The enhancement is more important for the prediction-based technique because of the conservative Bonferroni correction. The sensitivity is also improved for \textit{Test\_X}, where the correction is conservative because of features correlation. This adaptation of the significance level reduces the gap between the two approaches, bringing closer prediction-based and feature-based shift detectors. 

As an ablation study we investigate the improvements of the detectors ensembles due to the adaptation of the significance level and confirm the benefit of this technique as shown in Table \ref{tab:adapt-avg-ranks-ensemble}.

 

\begin{table}[h]
\caption{Average ranks of $3$ shift detectors based on efficiency score, comparing fixed and adaptive significance levels. The best values between regular and adaptive versions of the same shift detector are in bold, while the best values overall are underlined.}
\label{tab:adapt-avg-ranks}
\begin{adjustbox}{width=0.85\columnwidth,center}
\begin{tabular}{lcccccc}
\toprule
Shift type & BBSDs & BBSDs (adapt) & DC & DC (adapt) & Test\_X & Test\_X (adapt)\\
\midrule
Knock-out                          & 3.17 &\underline{\textbf{2.12}} & 4.43 & 4.43 & 4.02 & \textbf{2.83} \\
Only-One                          & 3.10 & \underline{\textbf{2.29}} & 4.50 & \textbf{4.36} & 3.95 & \textbf{2.81} \\
Small Gaussian                    & 5.29 & \textbf{3.67} & 3.10 & \textbf{2.64} & 3.98 & \underline{\textbf{2.33}} \\
Medium Gaussian                  & 4.38 & \textbf{2.93} & 3.76 & \textbf{3.40} & 4.31 & \underline{\textbf{2.21}} \\
Adv. ZOO      & 4.40 & \textbf{3.64} & 3.76 & \textbf{3.64} & 3.10 & \underline{\textbf{2.45}} \\
Adv. Boundary & 4.02 & \textbf{2.83} & 4.38 & \textbf{4.24} & 3.64 &\underline{\textbf{1.88}} \\
Joint Subsampling             & 4.02 & \textbf{3.50} & 4.02 & \textbf{3.86} & 3.29 & \underline{\textbf{2.31}} \\
Subsampling           & 4.60 & \textbf{3.98} & 3.81 & \textbf{3.57} & 3.19 & \underline{\textbf{1.86}} \\
Under-sampling               & 5.38 & \textbf{4.50} & 3.17 & 3.17 & 2.81 & \underline{\textbf{1.98}} \\
Over-sampling                 & 5.67 & \textbf{4.60} & 3.02 & \textbf{2.69} & 2.81 & \underline{\textbf{2.21}} \\
\bottomrule
\end{tabular}
\end{adjustbox}
\end{table}
 
\begin{table}[h]
\caption{Average ranks of $3$ ensemble shift detectors based on efficiency score, comparing fixed and adaptive significance levels. The best values between regular and adaptive versions of the same shift detector are in bold, while the best values overall are underlined.}
\label{tab:adapt-avg-ranks-ensemble}
\begin{adjustbox}{width=\columnwidth,center}
\begin{tabular}{lcccccc}
\toprule
Shift type & BBSDs+X & BBSDs+X (adapt) & BBSDs+DC & BBSDs+DC (adapt) & DC* & DC* (adapt)\\
\midrule
Knock-out                          & 3.55 & \underline{\textbf{1.95}} & 3.43 & \textbf{2.93} & 4.57 & 4.57 \\
Only-One                          & 3.55 & \underline{\textbf{2.62}} & 3.40 & \textbf{2.83} & 4.36 & \textbf{4.24} \\
Small Gaussian                 & 4.29 & \underline{\textbf{2.26}} & 3.40 & \textbf{2.62} & 4.36 & \textbf{4.07} \\
Medium Gaussian            & 4.19 & \underline{\textbf{2.21}} & 3.98 & \textbf{2.24} & 4.50 & \textbf{3.88} \\
Adv. ZOO                         & 3.24 & \underline{\textbf{2.57}} & 3.81 & \textbf{3.52} & 3.85 & 3.85 \\
Adv. Boundary                 & 3.21 & \underline{\textbf{1.88}} & 3.86 & \textbf{3.17} & 4.62 & \textbf{4.15} \\
Joint Subsampling            & 3.24 & \underline{\textbf{2.43}} & 3.79 & \textbf{3.55} & 4.00 & 4.00 \\
Subsampling                    & 2.67 & \underline{\textbf{2.19}} & 3.31 & \textbf{2.79} & 5.02 & 5.02 \\
Under-sampling               & 3.43 & \underline{\textbf{2.57}} & 4.00 & \textbf{3.57} & 3.71 & 3.71 \\
Over-sampling                 & 3.60 & \underline{\textbf{2.88}} & 3.88 & \textbf{3.17} & 3.65 & 3.65 \\
\bottomrule
\end{tabular}
\end{adjustbox}
\end{table}

\subsection{Impact of Model Quality on BBSDs}
\label{subsec:res_quality}

To confirm the limitations of BBSDs on the primary model performance  (cf. Section \ref{sec:dd}), we apply a random perturbation to the primary model with probability $p$, yielding a primary model with quality $1-p$ in $[0.5, 1.0]$ and evaluate the power of the BBSDs test. As illustrated in Figure \ref{fig:quality-boxplot}, showing the average TPR across $21$ datasets with size of $1000$ samples, the power decreases with the primary model quality, along with the variance of the results across the datasets. More difficult shifts (such as low intensity or a small amount of drifted samples) require a better primary model for the BBSDs test power to hold (shift-specific details in Table \ref{tab:p-val-shift} of the Appendix). Regardless of the model quality, BBSDs is also less adequate in detecting some types of selection bias, as already observed in \mbox{Figure \ref{fig:tpr-boxplots}}.
 
\section{Conclusion}

In this paper we propose a shift detectors ensembling technique capable of addressing all different types of studied drift scenarios. 
The key components of the proposed approach are the combination of complementary base detectors, designed to address different types of shift, and the adaptation of the significance level of the detectors statistical tests to the specific dataset under study. 

The improved robustness of our approach is validated by a large-scale benchmark study comparing it to state-of-the-art shift detectors on $21$ structured real-world datasets. For this purpose we simulate drifts with $10$ different types of perturbations, including shift types not studied in previous works for tabular datasets, such as adversarial noise and sample selection bias. Our benchmark study highlights the complementarity of base drift detectors on different drift scenarios, motivating our ensembling approach. This benchmark also includes ablation studies showing that the dataset-adaptive significance level provides both base and ensembled shift detectors with higher detection power, while preserving the desired false positive rate. 

Throughout our experiments, we observe that adaptive shift detectors ensembling represents the strongest strategy, robust to the various shift types, making this approach a natural choice for monitoring models in production in real-life settings, where the possible drift scenario is unknown.\\


\appendix
\section{Appendix}
\label{sec:app}

\subsection{Proof from Subsection \ref{subsec:adapt-sign-level}}
\label{subsec:app-bbsd}

\begin{proof}[There exist prior shifts such that the smaller the predictive power of the model is, the higher the $p$-values of the BBSD test are]
The true and predicted class distributions in the source (resp. target) domain are denoted by $\mathbf{p}_S$ and $\hat{\mathbf{p}}_S$ (resp. $\mathbf{p}_T$ and $\hat{\mathbf{p}}_T$).

Let us take a model $\mathit{f}: \mathit{X} \rightarrow \mathit{Y}$ solving a classification task with an ill-conditioned but still invertible confusion matrix $\mathbf{C}$, with eigenvalues $\lambda_i$ and eigenvectors $\mathbf{v}_i$.  Its minimum eigenvalue in absolute value is $\lambda_{min}$ and the corresponding eigenvector is $\mathbf{v}_{min}$. 

First, we make clear the connection between the model's predictive performance and the norm of the minimum eigenvalue. We perturb the primary model $\mathit{f}$ prediction's by predicting random labels with probability $p$. The new model is denoted by $\tilde{f}$. As the perturbation $p$ tends to $1$, the eigenvalue $\lambda_{min}(\tilde{f})$ tends to $0$. We then consider the family of prior shift $\mathbf{p}_{T} = \mathbf{p}_{S} + \alpha \cdot \mathbf{v}_{min}(\tilde{f})$ indexed by $\alpha$ such that $\mathbf{p}_{T}$ is a probability measure on $Y$.

By Lemma 1 of \cite{lipton18} under the prior shift assumption and by the law of total probability we have that the following equation holds for both the source and the target domain:
\begin{align*}
\mathbb{P}(\hat y) = \sum_{y \in \mathcal{Y} } \mathbb{P}(\hat y | y)  \mathbb{P}(y)
\end{align*}

Considering the column-normalized confusion matrix $\mathbf{C}$ as an estimator for $\mathbb{P}(\hat y | y)$, we can write in matrix form:

\begin{align*}
\mathbf{\hat p}_S=\mathbf{C} \cdot \mathbf{p}_S \qquad \mathbf{\hat p}_T=\mathbf{C} \cdot \mathbf{p}_T 
\end{align*}

Let us assume $\mathbf{p}_S\neq\mathbf{p}_T$ with $\mathbf{p}_T=\mathbf{p}_S + \alpha \cdot \mathbf{v}_{min}$ with $\alpha \in [\alpha_{min},\alpha_{max}]$, such that $p_{Ti} \in [0, 1] \forall i$. Note that $\alpha_{min}$ is always non positive and $\alpha_{max}$ is always non negative. 

For the target distribution we have that:
\begin{align*}
\mathbf{\hat p}_T&=\mathbf{C} \cdot \mathbf{p}_T = \mathbf{C} \cdot \mathbf{p} _S+ \mathbf{C} \cdot \alpha \cdot \mathbf{v}_{min}\\
&= \mathbf{\hat p}_S + \alpha \cdot \lambda_{min} \cdot \mathbf{v}_{min} \approx \mathbf{\hat p}_S
\end{align*}
Thus $\mathbf{\hat p}_S \approx \mathbf{\hat p}_T$ while $\mathbf{p}_S \neq \mathbf{p}_T$.

In particular:
\begin{align*}
\norm{\mathbf{\hat p} _S- \mathbf{\hat p}_T}_2 = \vert\alpha \cdot \lambda_{min}\vert \approx 0
\end{align*}
\begin{align*}
\norm{\mathbf{p}_S - \mathbf{p}_T}_2 = \vert\alpha\vert
\end{align*}

with $ \vert\alpha\vert  \in [0,\max\{ \vert\alpha_{min}\vert, \alpha_{max}\}]$.

This shows that as the perturbation $p$ tends to $1$, the $\ell_{2}$ norm $\norm{ \mathbf{\hat p}_S - \mathbf{\hat p}_T}_{2}$ tends to $0$ while $\norm{ \mathbf{p}_S - \mathbf{p}_T}_{2} = \alpha$. As the $\chi^2$ statistic is controlled by $\ell_{2}\text{-norm}$, this means BBSD power is limited by the strength of the perturbation.
\end{proof}

\subsection{Experiments Details}
\label{subsec:app-exp}

\subsubsection{Datasets}
Details on the $21$ OpenML\footnote{www.openml.org} and Kaggle \footnote{www.kaggle.com} structured classification datasets are available in Table \ref{tab:datasets}.

\begin{table}[t]
\caption{Classification datasets from OpenML and Kaggle (denoted by a *).}
\label{tab:datasets}
\begin{adjustbox}{width=0.7\columnwidth,center}
\begin{tabular}{lcccc}
\toprule
Dataset & classes & features & size & \% minority \\
&  &  &  & class \\
\midrule
adult                     & 2 & 14 & 48842 & 23.9 \\
Amazon employee access & 2 & 9 & 32769 & 5.8  \\
Artificial characters  & 10 & 7 & 10218 & 5.9\\
Click prediction small & 2 & 9 & 39948 & 16.8\\
codrna                  & 2 & 8 & 488565 & 33.3\\
creditcard             & 2 & 30 & 284807 & 0.17\\
electricity              & 2 & 8 & 45312 & 42.5\\
higgs                     & 2 & 29 & 98050 & 47.1\\
homesite-quote-conversion* & 2 & 298 & 260753 & 18.8 \\
JapaneseVowels  & 9 & 14 & 9961 & 7.9\\
jm1                        & 2 & 21 & 10885 & 19.3\\
letter                     & 26 & 16 & 20000 & 3.7\\
MagicTelescope   & 2 & 10 & 19020 & 35.2\\
mc1                       & 2 & 38 & 9466 & 0.72\\
mozilla4                & 2 & 5  & 15545 & 32.9\\
optdigits               & 10 & 64 & 5620 & 9.9\\
otto-group-product-classification* & 9 & 94 & 61878 & 3.1 \\
pc2                       & 2 & 36 & 5589 & 0.41\\
phoneme              & 2 & 5 & 5404 & 29.3\\ 
santander-customer-satisfaction* & 2 & 370 & 76020 & 4.0 \\ 
waveform\_5000  & 3 & 40 & 5000 & 33.1\\
\bottomrule
\end{tabular}
\end{adjustbox}
\end{table}

\subsubsection{Preprocessing and Algorithms Hyperparameters}

For each dataset (see Table \ref{tab:datasets}), a fixed training set of $1000$ samples is randomly selected. The remainder of each dataset is then randomly split into a validation set and test set of $2000$ samples each. The former is used as source dataset, while the latter is perturbed by applying a particular type of drift and constitutes the target dataset. Apart from simple imputation of missing values and one-hot encoding of categorical features, no preprocessing is performed. The training set is used to fit PCA and SRP parameters, as well as to train the primary model, a Random Forest, with default \textit{sklearn} hyper-parameters. Random Forest is an industry standard model, extensively employed in enterprise machine learning  for its performance and interpretability. The number of dimensions of PCA- and SRP-processed features is set to the same value corresponding to a variance retention rate in PCA of $0.8$.
The validation-test split is repeated over 5 random seeds so that each drift detection experiment is repeated 5 times.

\subsubsection{Reproducibility}

We leveraged various open source librairies to generate those shifts in our experiments: the prior and Gaussian shifts rely on the library provided in the Failing Loudly github repository\footnote{\url{https://github.com/steverab/failing-loudly}}, the adversarial shifts are generated using the Adversarial Robustness Toolbox\footnote{\url{https://github.com/IBM/adversarial-robustness-toolbox}}, while the under-sampling and over-sampling shifts rely on techniques from imbalanced-learn\footnote{\url{https://github.com/scikit-learn-contrib/imbalanced-learn}}. For the sake of completeness and reproducibility, all the code used to generate the shifts and to perform drift experiments is accessible in a public repository\footnote{\url{https://github.com/dataiku-research/drift_detectors_benchmark}}.

\subsection{Detailed Results from Subsections \ref{subsec:res_bench}, \ref{subsec:impact_adap} and  \ref{subsec:res_quality}}
Values are reported with $mean \pm std$. Best values and all values within $95$\% confidence interval are in bold.

\begin{table}[h]
\caption{Accuracy across all datasets by size.}
\label{tab:acc-size}
\begin{adjustbox}{width=0.9\columnwidth,center}
\begin{tabular}{lcccccc}
\toprule
Size  &  BBSDh &  BBSDs &    DC &  Test\_PCA &  Test\_SRP &  Test\_X \\
\midrule
 10   & 0.10 \tiny{$\pm$ 0.02} & 0.11 \tiny{$\pm$ 0.03} & 0.10 \tiny{$\pm$ 0.02} & 0.13 \tiny{$\pm$ 0.03} & 0.11 \tiny{$\pm$ 0.03} & 0.10 \tiny{$\pm$ 0.02} \\
 100  & 0.15 \tiny{$\pm$ 0.06} & 0.30 \tiny{$\pm$ 0.10} & \textbf{0.45} \tiny{$\pm$ 0.09} & 0.36 \tiny{$\pm$ 0.06} & 0.29 \tiny{$\pm$ 0.10} & \textbf{0.45} \tiny{$\pm$ 0.10} \\
 500  & 0.22 \tiny{$\pm$ 0.11} & 0.46 \tiny{$\pm$ 0.14} & \textbf{0.64} \tiny{$\pm$ 0.10} & 0.47 \tiny{$\pm$ 0.11} & 0.47 \tiny{$\pm$ 0.14} & \textbf{0.65} \tiny{$\pm$ 0.12} \\
 1000 & 0.27 \tiny{$\pm$ 0.15} & 0.54 \tiny{$\pm$ 0.16} & \textbf{0.66} \tiny{$\pm$ 0.11} & 0.51 \tiny{$\pm$ 0.13} & 0.54 \tiny{$\pm$ 0.16} & \textbf{0.69} \tiny{$\pm$ 0.11} \\
 2000 & 0.33 \tiny{$\pm$ 0.19} & 0.63 \tiny{$\pm$ 0.19} & \textbf{0.71} \tiny{$\pm$ 0.14} & 0.55 \tiny{$\pm$ 0.20} & 0.60 \tiny{$\pm$ 0.19} & \textbf{0.75} \tiny{$\pm$ 0.13} \\
\bottomrule
\end{tabular}
\end{adjustbox}
\end{table}

\begin{table}[h]
\caption{TPR across all datasets by shift type at dataset size of $1000$.}
\label{tab:tpr-shift}
\begin{adjustbox}{width=0.9\columnwidth,center}
\begin{tabular}{lcccccc}
\toprule
Shift type &  BBSDh &  BBSDs &    DC &  Test\_PCA &  Test\_SRP &  Test\_X \\
\midrule
 Knock-Out         & \textbf{0.15} \tiny{$\pm$ 0.27} & \textbf{0.23} \tiny{$\pm$ 0.30} & 0.00 \tiny{$\pm$ 0.00} & 0.06 \tiny{$\pm$ 0.18} & 0.09 \tiny{$\pm$ 0.21} & \textbf{0.11} \tiny{$\pm$ 0.24} \\
 Only-One          & \textbf{0.50} \tiny{$\pm$ 0.52} & \textbf{0.60} \tiny{$\pm$ 0.46} & \textbf{0.46} \tiny{$\pm$ 0.47} & 0.36 \tiny{$\pm$ 0.43} & \textbf{0.47} \tiny{$\pm$ 0.48} & \textbf{0.55} \tiny{$\pm$ 0.49} \\
 Small Gaussian    & 0.04 \tiny{$\pm$ 0.14} & 0.74 \tiny{$\pm$ 0.37} & \textbf{1.00} \tiny{$\pm$ 0.00} & 0.24 \tiny{$\pm$ 0.33} & 0.41 \tiny{$\pm$ 0.45} & 0.96 \tiny{$\pm$ 0.14} \\
 Medium Gaussian   & 0.28 \tiny{$\pm$ 0.33} & 0.87 \tiny{$\pm$ 0.32} & \textbf{1.00} \tiny{$\pm$ 0.00} & 0.49 \tiny{$\pm$ 0.45} & 0.66 \tiny{$\pm$ 0.45} & 0.98 \tiny{$\pm$ 0.06} \\
 Adv. ZOO          & 0.14 \tiny{$\pm$ 0.36} & \textbf{0.21} \tiny{$\pm$ 0.40} & \textbf{0.42} \tiny{$\pm$ 0.50} & \textbf{0.22} \tiny{$\pm$ 0.40} & \textbf{0.32} \tiny{$\pm$ 0.47} & \textbf{0.43} \tiny{$\pm$ 0.51} \\
 Adv. Boundary     & 0.47 \tiny{$\pm$ 0.45} & \textbf{0.72} \tiny{$\pm$ 0.42} & \textbf{0.70} \tiny{$\pm$ 0.41} & 0.33 \tiny{$\pm$ 0.38} & 0.53 \tiny{$\pm$ 0.45} & \textbf{0.77} \tiny{$\pm$ 0.35} \\
 Subsampling Joint & 0.00 \tiny{$\pm$ 0.00} & 0.04 \tiny{$\pm$ 0.08} & 0.10 \tiny{$\pm$ 0.19} & \textbf{0.38} \tiny{$\pm$ 0.39} & \textbf{0.20} \tiny{$\pm$ 0.28} & \textbf{0.20} \tiny{$\pm$ 0.30} \\
 Subsampling       & 0.29 \tiny{$\pm$ 0.35} & 0.42 \tiny{$\pm$ 0.41} & \textbf{0.70} \tiny{$\pm$ 0.45} & \textbf{0.66} \tiny{$\pm$ 0.41} & \textbf{0.70} \tiny{$\pm$ 0.41} & \textbf{0.71} \tiny{$\pm$ 0.45} \\
 Under-sampling    & 0.04 \tiny{$\pm$ 0.13} & 0.45 \tiny{$\pm$ 0.45} & \textbf{1.00} \tiny{$\pm$ 0.00} & \textbf{1.00} \tiny{$\pm$ 0.00} & 0.74 \tiny{$\pm$ 0.45} & \textbf{1.00} \tiny{$\pm$ 0.00} \\
 Over-sampling     & 0.02 \tiny{$\pm$ 0.06} & 0.57 \tiny{$\pm$ 0.40} & \textbf{1.00} \tiny{$\pm$ 0.00} & \textbf{1.00} \tiny{$\pm$ 0.00} & 0.81 \tiny{$\pm$ 0.35} & \textbf{1.00} \tiny{$\pm$ 0.00} \\
\bottomrule
\end{tabular}
\end{adjustbox}
\end{table}

\begin{table}[h!]
\caption{BBSDs TPR across all datasets by quality and shift at $1000$ samples.}
\label{tab:p-val-shift}
\begin{adjustbox}{width=0.9\columnwidth,center}
\begin{tabular}{lcccccc}
\toprule
Quality  &  0.5 &   0.6 &   0.7  &   0.8  &   0.9  &   1.0  \\
(mean accuracy)  &    (40\%)   &   (44\%)     &   (52\%)    &    (63\%)   &  (75\%)     &   (87\%)    \\
Shift type &   &   &     &   &   &   \\
\midrule
 Knock-Out         & 0.00 \tiny{$\pm$ 0.00} & 0.03 \tiny{$\pm$ 0.07} & 0.13 \tiny{$\pm$ 0.09} & 0.05 \tiny{$\pm$ 0.08} & \textbf{0.15} \tiny{$\pm$ 0.23} & \textbf{0.28} \tiny{$\pm$ 0.31} \\
 Only-One          & 0.00 \tiny{$\pm$ 0.00} & 0.08 \tiny{$\pm$ 0.10} & \textbf{0.39} \tiny{$\pm$ 0.29} & \textbf{0.51} \tiny{$\pm$ 0.49} & \textbf{0.57} \tiny{$\pm$ 0.50} & \textbf{0.59} \tiny{$\pm$ 0.49} \\
 Small Gaussian    & 0.00 \tiny{$\pm$ 0.00} & 0.05 \tiny{$\pm$ 0.07} & 0.42 \tiny{$\pm$ 0.27} & \textbf{0.70} \tiny{$\pm$ 0.34} & \textbf{0.76} \tiny{$\pm$ 0.35} & \textbf{0.82} \tiny{$\pm$ 0.34} \\
 Medium Gaussian   & 0.00 \tiny{$\pm$ 0.00} & 0.10 \tiny{$\pm$ 0.12} & 0.68 \tiny{$\pm$ 0.29} & \textbf{0.88} \tiny{$\pm$ 0.30} & \textbf{0.88} \tiny{$\pm$ 0.31} & \textbf{0.95} \tiny{$\pm$ 0.18} \\
 Adv. ZOO          & 0.00 \tiny{$\pm$ 0.00} & 0.03 \tiny{$\pm$ 0.07} & \textbf{0.17} \tiny{$\pm$ 0.15} & \textbf{0.15} \tiny{$\pm$ 0.31} & \textbf{0.19} \tiny{$\pm$ 0.35} & \textbf{0.21} \tiny{$\pm$ 0.39} \\
 Adv. Boundary     & 0.00 \tiny{$\pm$ 0.00} & 0.02 \tiny{$\pm$ 0.04} & 0.33 \tiny{$\pm$ 0.19} & \textbf{0.57} \tiny{$\pm$ 0.36} & \textbf{0.67} \tiny{$\pm$ 0.36} & \textbf{0.73} \tiny{$\pm$ 0.37} \\
 Subsampling Joint & 0.00 \tiny{$\pm$ 0.00} & 0.03 \tiny{$\pm$ 0.07} & \textbf{0.13} \tiny{$\pm$ 0.10} & 0.03 \tiny{$\pm$ 0.07} & 0.00 \tiny{$\pm$ 0.00} & 0.04 \tiny{$\pm$ 0.08} \\
 Subsampling       & 0.00 \tiny{$\pm$ 0.00} & 0.00 \tiny{$\pm$ 0.00} & \textbf{0.13} \tiny{$\pm$ 0.10} & 0.03 \tiny{$\pm$ 0.07} & 0.04 \tiny{$\pm$ 0.10} & \textbf{0.10} \tiny{$\pm$ 0.15} \\
 Under-sampling    & 0.00 \tiny{$\pm$ 0.00} & 0.04 \tiny{$\pm$ 0.08} & 0.14 \tiny{$\pm$ 0.11} & 0.18 \tiny{$\pm$ 0.32} & 0.20 \tiny{$\pm$ 0.34} & \textbf{0.45} \tiny{$\pm$ 0.45} \\
 Over-sampling     & 0.00 \tiny{$\pm$ 0.00} & 0.04 \tiny{$\pm$ 0.08} & 0.16 \tiny{$\pm$ 0.10} & 0.23 \tiny{$\pm$ 0.35} & \textbf{0.36} \tiny{$\pm$ 0.38} & \textbf{0.52} \tiny{$\pm$ 0.42} \\
\bottomrule
\end{tabular}
\end{adjustbox}
\end{table}

 \begin{figure*}[t!]
\centering
\subfloat[Knock-Out shift]{
  \includegraphics[width=.45\linewidth]{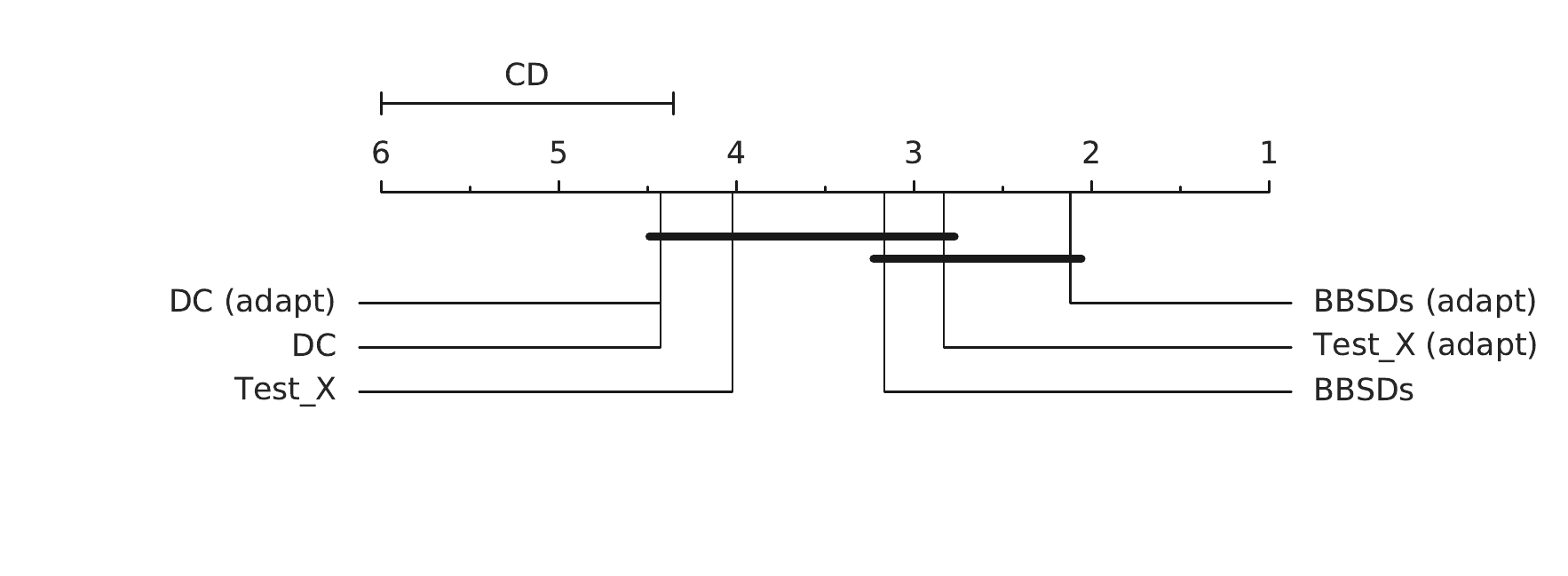}  
  \label{fig:ko_ad}
}
\subfloat[Only-One shift]{
  \includegraphics[width=.45\linewidth]{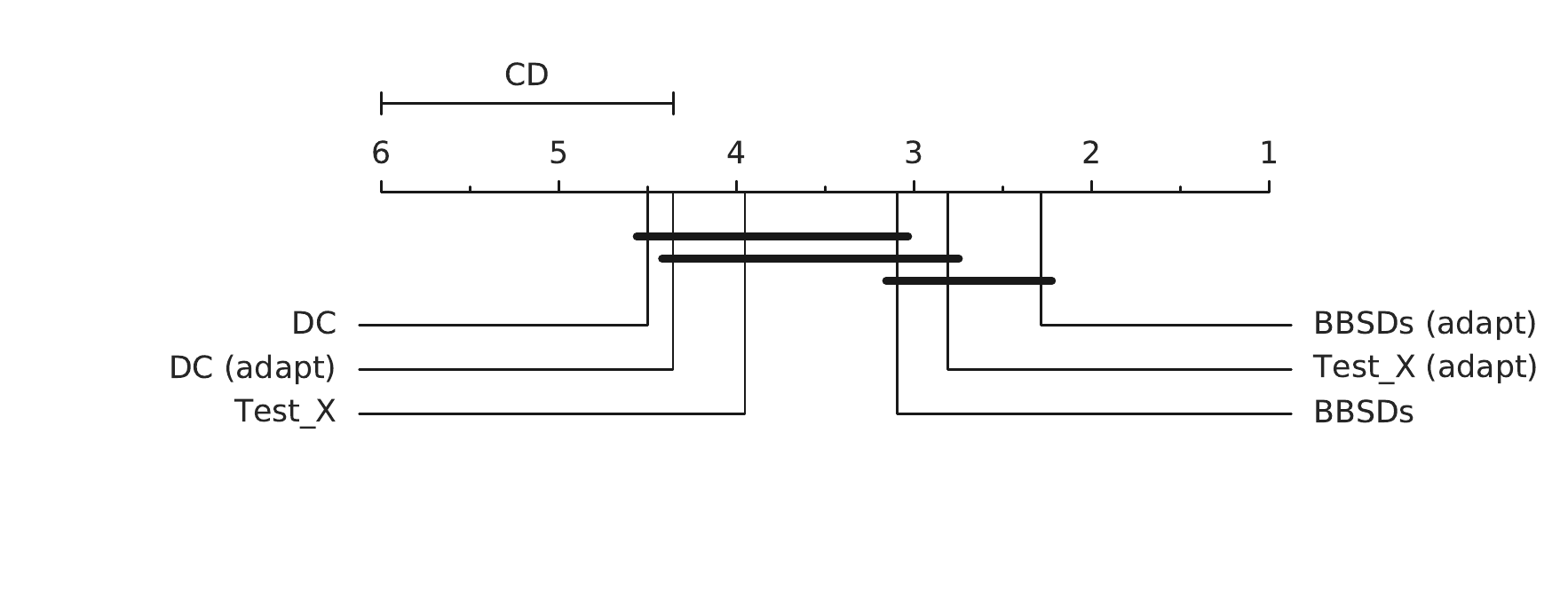}  
  \label{fig:oo_ad}
}\\
\subfloat[Small Gaussian Noise shift]{
  \includegraphics[width=.45\linewidth]{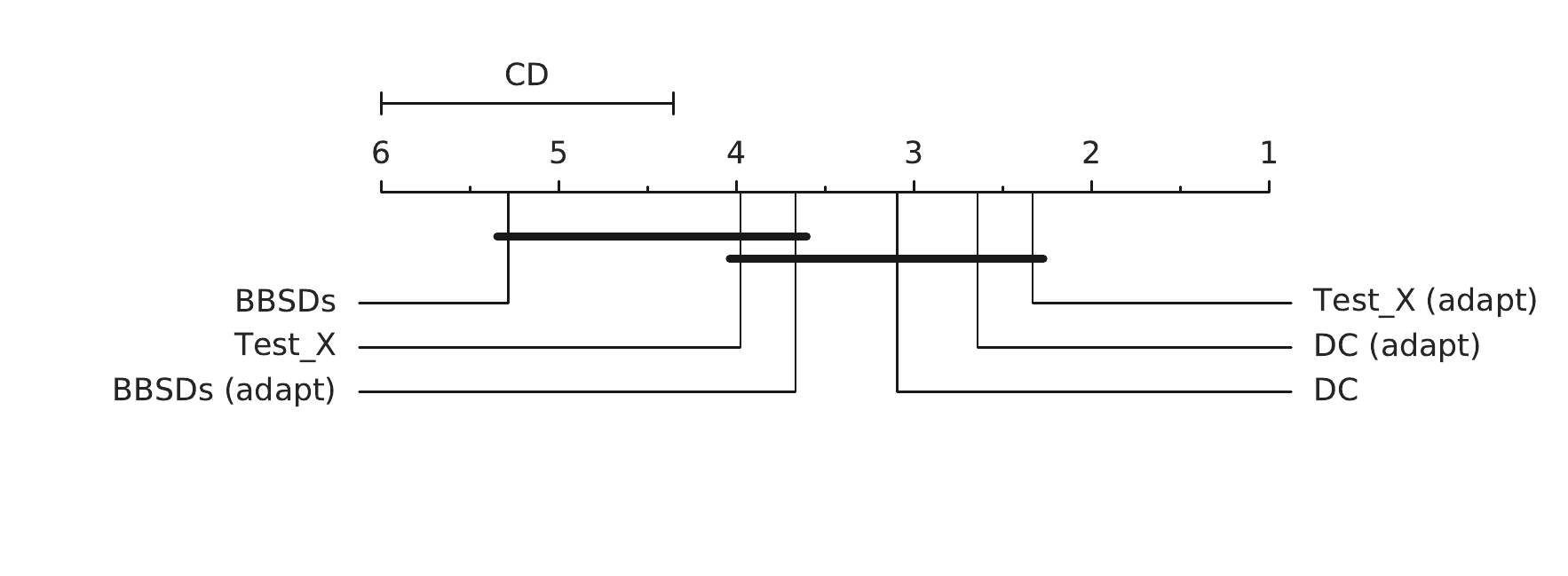}  
  \label{fig:sg_ad}
}
\subfloat[Medium Gaussian Noise shift]{
  \includegraphics[width=.45\linewidth]{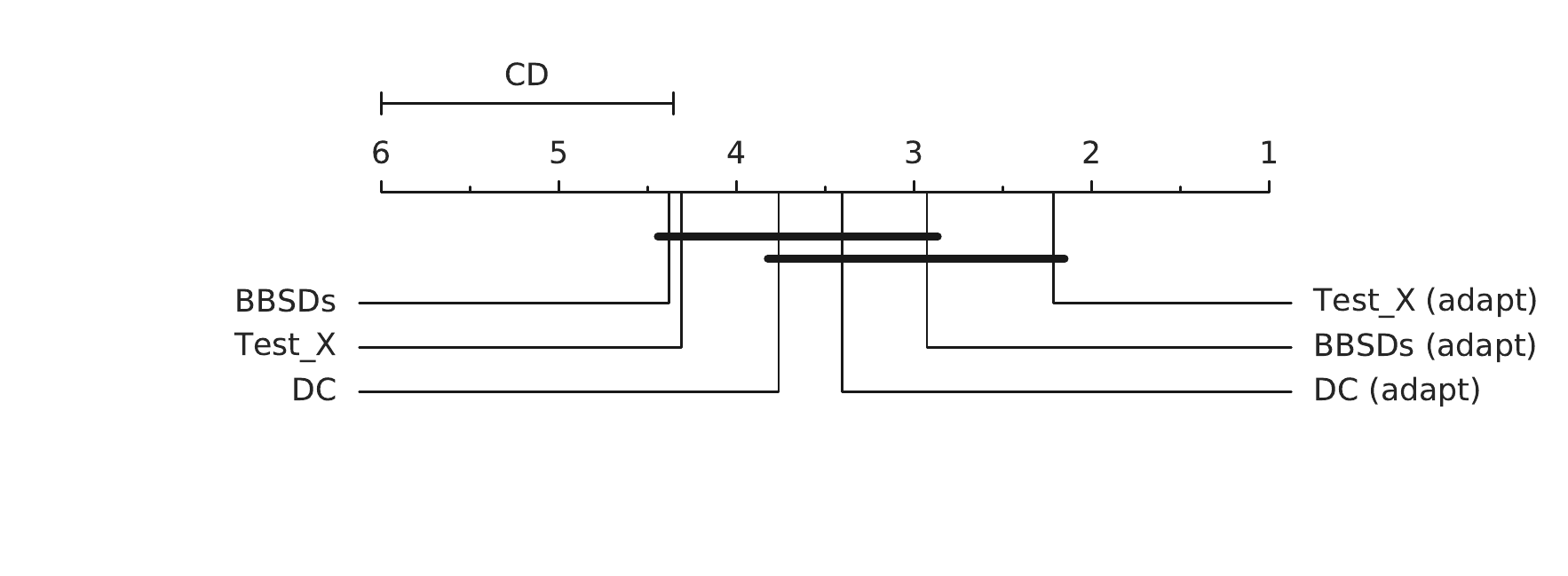}  
  \label{fig:mg_ad}
}\\
\subfloat[Adversarial ZOO perturbation]{
  \includegraphics[width=.45\linewidth]{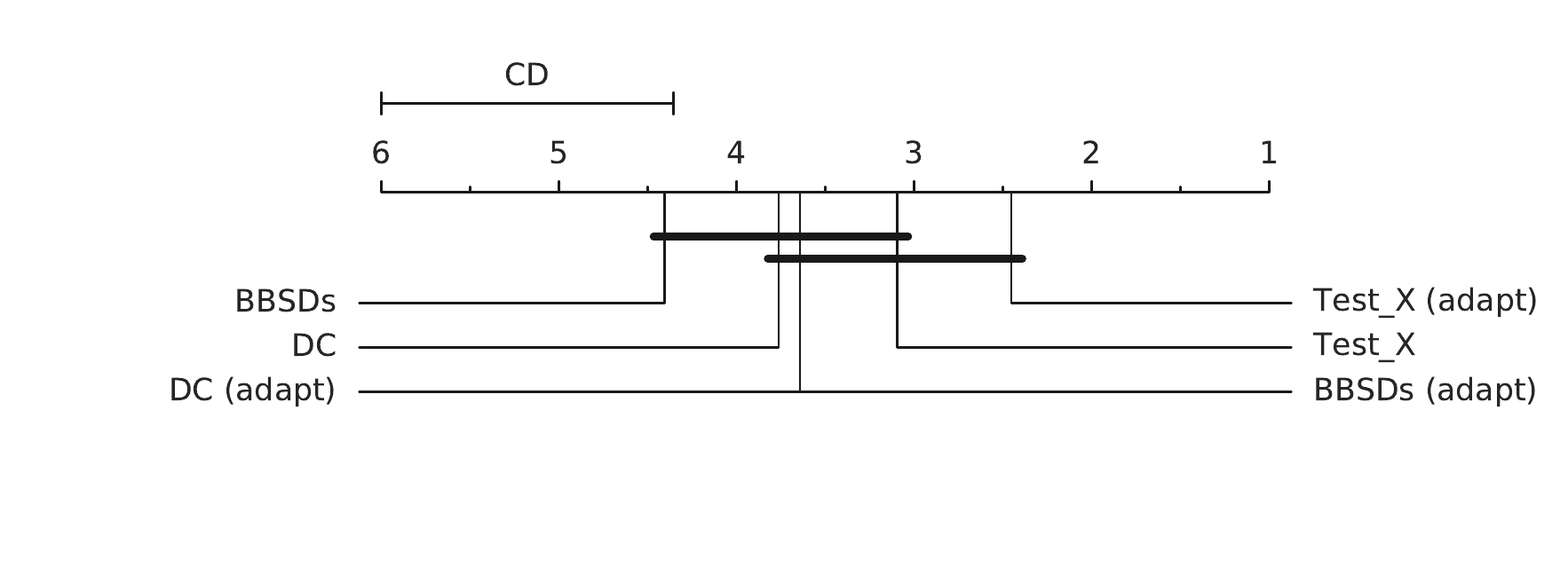}  
  \label{fig:advz_ad}
}
\subfloat[Adversarial Boundary perturbation]{
  \includegraphics[width=.45\linewidth]{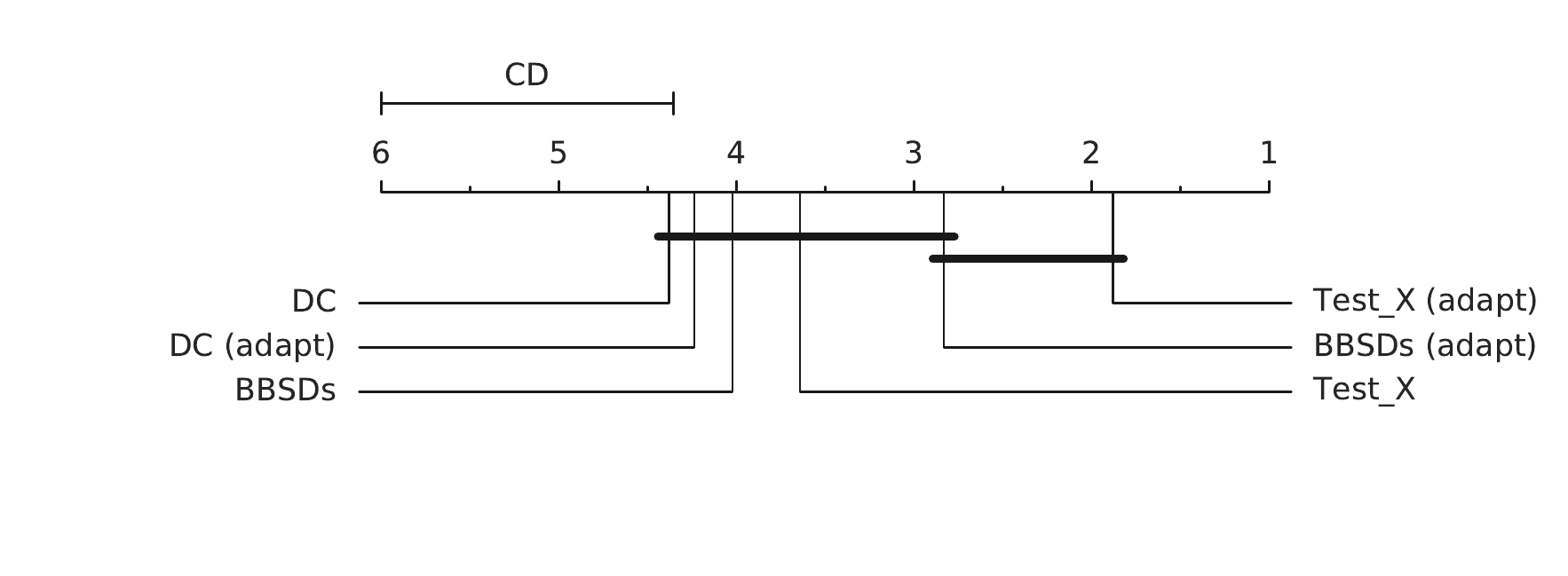}  
  \label{fig:advb_ad}
}\\
\subfloat[Joint Subsampling shift]{
  \includegraphics[width=.45\linewidth]{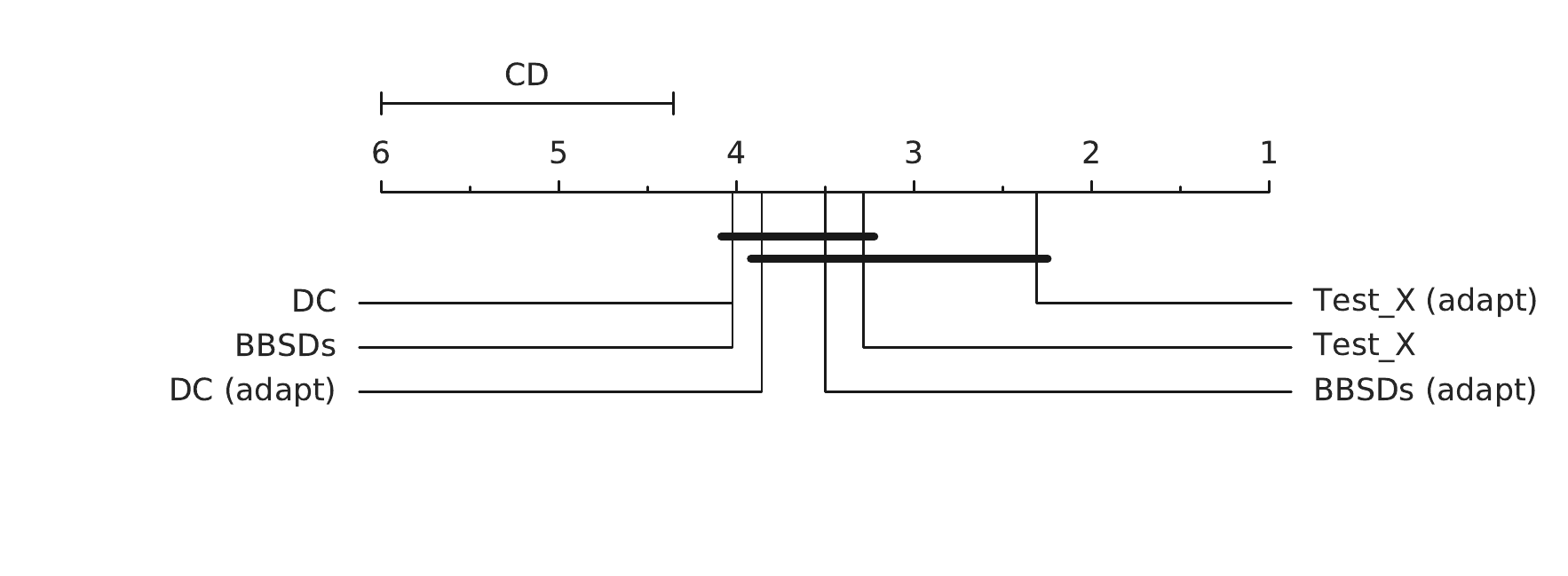}  
  \label{fig:js_ad}
}
\subfloat[Subsampling shift]{
  \includegraphics[width=.45\linewidth]{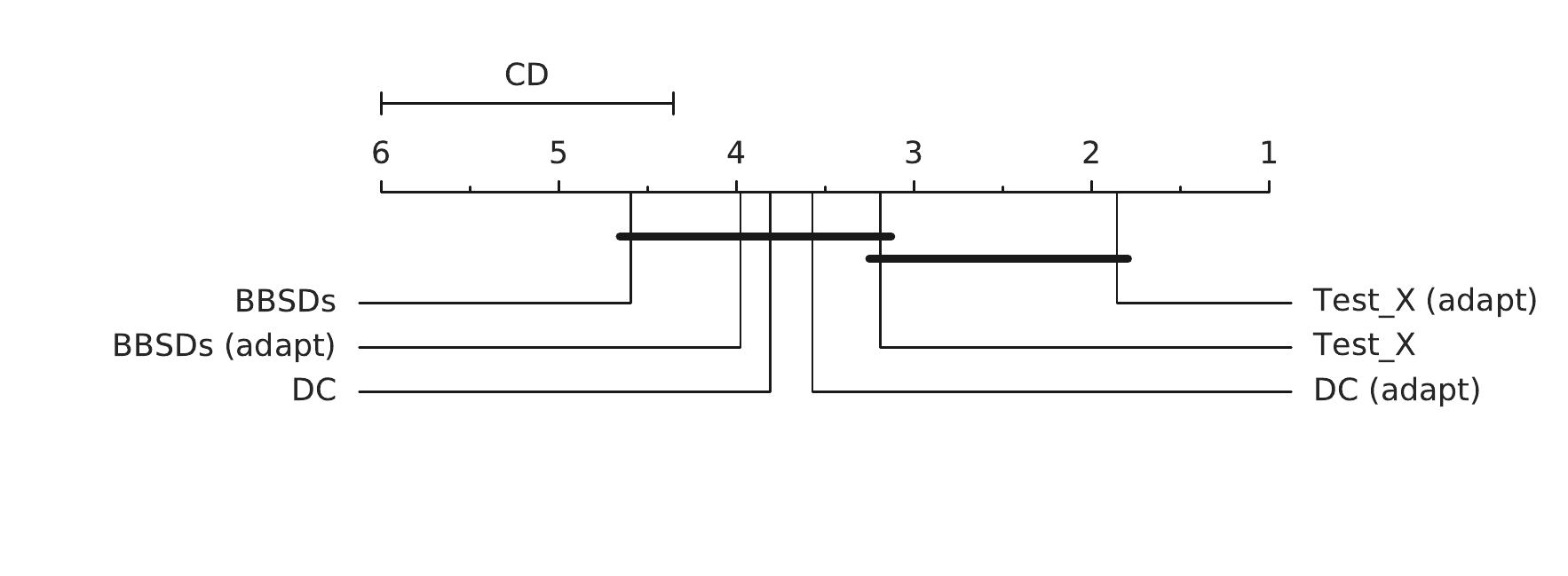}  
  \label{fig:ss_ad}
}\\
\subfloat[Under-sampling shift]{
  \includegraphics[width=.45\linewidth]{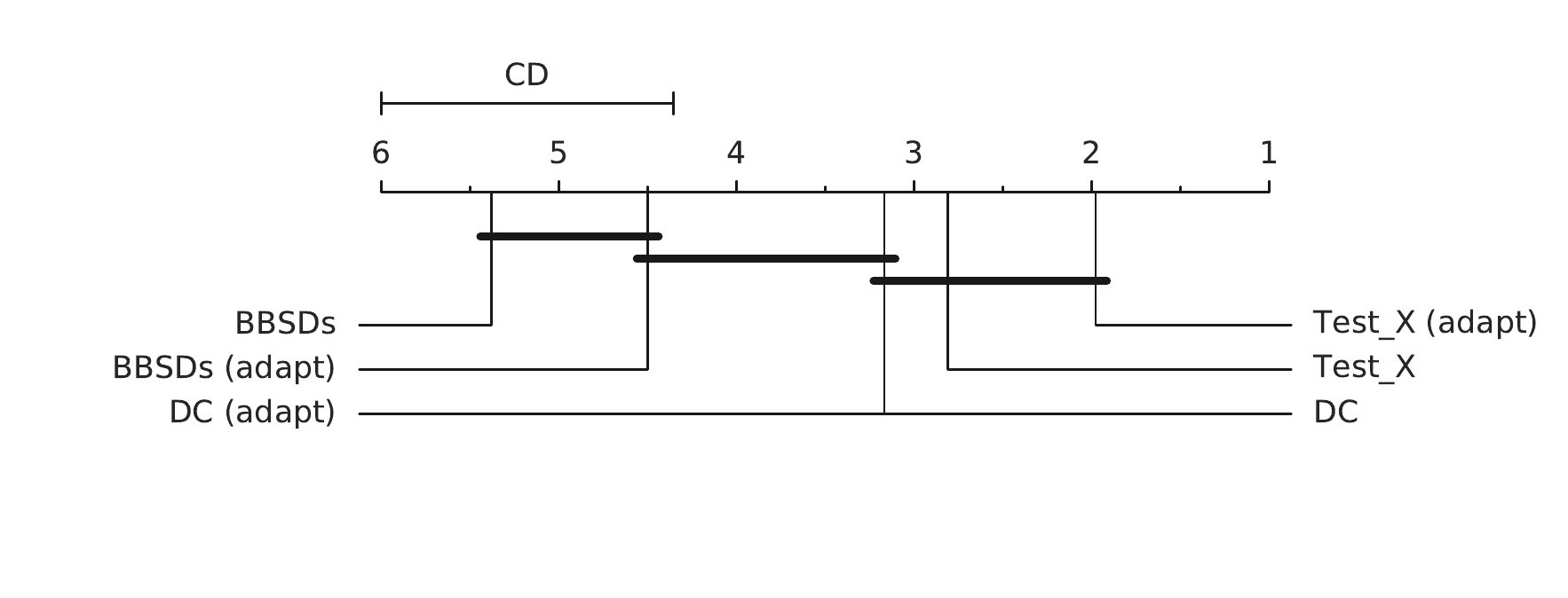}  
  \label{fig:us_ad}
}
\subfloat[Over-sampling shift]{
  \includegraphics[width=.45\linewidth]{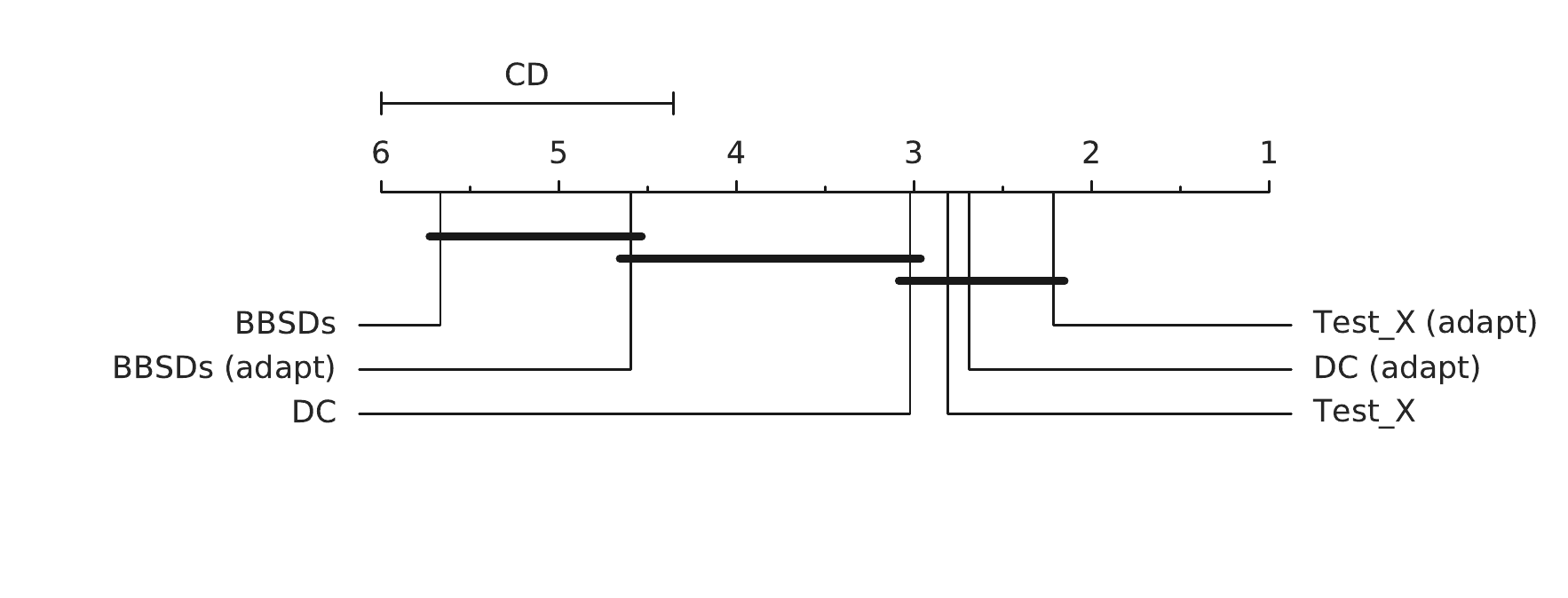}  
  \label{fig:os_ad}
}
\caption{Nemenyi post-hoc test comparing regular and adaptive significance level.}
\label{fig:adapt-nemenyi-plots}
\vskip -0.2in
\end{figure*}

\bibliographystyle{splncs04}

\end{document}